\pdfoutput=1

\documentclass[10pt,twocolumn,letterpaper]{article}

\usepackage[pagenumbers]{cvpr} 

\usepackage[dvipsnames]{xcolor}

\usepackage{graphicx}
\usepackage{amsmath}
\usepackage{amssymb}
\usepackage{booktabs}
\usepackage{todonotes}
\usepackage{tabularx}
\usepackage{multirow}
\usepackage{multicol}
\usepackage{bm}
\usepackage{url}
\usepackage{balance}
\usepackage{enumitem}
\usepackage{color}
\usepackage{bbding}
\usepackage{nccmath}
\usepackage{adjustbox}
\usepackage{tikzinclude}

\definecolor{cvprblue}{rgb}{0.21,0.49,0.74}
\usepackage[pagebackref,breaklinks,colorlinks,allcolors=cvprblue]{hyperref}

\usepackage{comment}
\usepackage[table]{colortbl}
\usepackage{dsfont}
\usepackage{etoolbox}
\usepackage{color}

\newif\ifshowedits

\newcommand{\addeditor}[3]{%
  \definecolor{#1color}{rgb}{#3}
  \expandafter\newcommand\csname #1\endcsname[1]{%
  \ifshowedits
    {\color{#1color} ##1}%
  \else
    {##1}%
  \fi
  }%
  \expandafter\newcommand\csname #1rmk\endcsname[1]{%
  \ifshowedits
    {\color{#1color} {\bf [#2: ##1]}}
  \fi
  }%
  \expandafter\newcommand\csname #1rpl\endcsname[2]{%
  \ifshowedits
    {\color{#1color} ##1 \sout{##2}}
  \else
    {##1}
  \fi
  }%
}

\newcommand\tocheck[1]{%
  \ifshowedits
    {\color{red} {\bf #1}}
  \fi
}

\newcommand{\createtextvar}[1]{
  \expandafter\newcommand\csname #1\endcsname{%
  {\text{#1}}
}%
}
\newcommand{\mycomment}[1]{}

\newcommand{\calL}{{\cal L}}

\newcommand{\IR}{{\mathds{R}}}

\newcommand{\vcomment}[1]{}

\usepackage{tikz}
\usetikzlibrary{arrows.meta, positioning, shadows.blur}

\addeditor{nishino}{NK}{0.0, 0.0, 0.8}
\addeditor{guedon}{GA}{0.0, 0.5, 0.0}
\addeditor{yamashita}{YK}{0.9, 0., 0.9}
\addeditor{kohei}{YK}{0.9, 0., 0.9} 
\addeditor{ichikawa}{IT}{0.85, 0.5, 0.0}
\showeditsfalse  

\newcommand{\method}{MAtCha Gaussians\xspace}
\newcommand{\shortmethod}{MAtCha\xspace}

\def\paperID{41} 
\def\confName{CVPR}
\def\confYear{2025}

\title{MAtCha Gaussians: Atlas of Charts for High-Quality Geometry and Photorealism From Sparse Views}

\author{
Antoine Guédon${}^{\dagger}$ 
\hspace{20pt} Tomoki Ichikawa${}^\ddagger$ 
\hspace{20pt} Kohei Yamashita${}^\ddagger$ 
\hspace{20pt} Ko Nishino${}^\ddagger$\\
$^\dagger$ LIGM, Ecole des Ponts, Univ Gustave Eiffel, CNRS, France \\
$^\ddagger$Graduate School of Informatics, Kyoto University, Japan \\ 
{\tt\small antoine.guedon@enpc.fr} \hspace{5pt} 
{\tt\small \{tichikawa,kyamashita\}@vision.ist.i.kyoto-u.ac.jp} \hspace{5pt} \\
{\tt\small kon@i.kyoto-u.ac.jp} \\
{\small \url{https://anttwo.github.io/matcha/}}
}

\begin{document}

\twocolumn[{
  \maketitle 
  \begin{center}
    \captionsetup{type=figure}
    \centering
    \begin{minipage}[b]{0.28\textwidth}
        \centering
        \includegraphics[trim={0 1.cm 0 0},clip,width=0.98\linewidth]{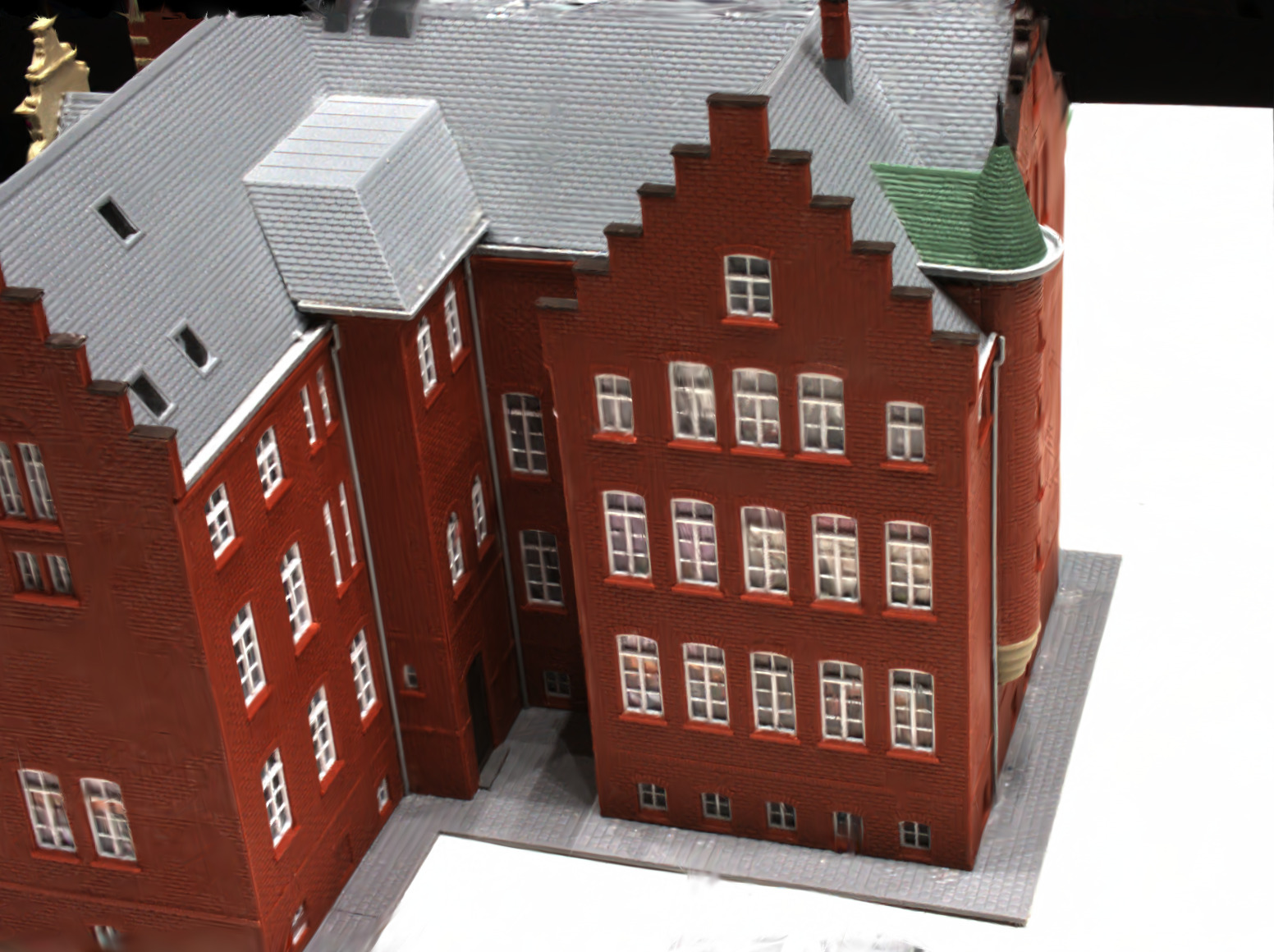}
        \vfill
        \includegraphics[trim={0 0 0 0},clip,width=0.98\linewidth]{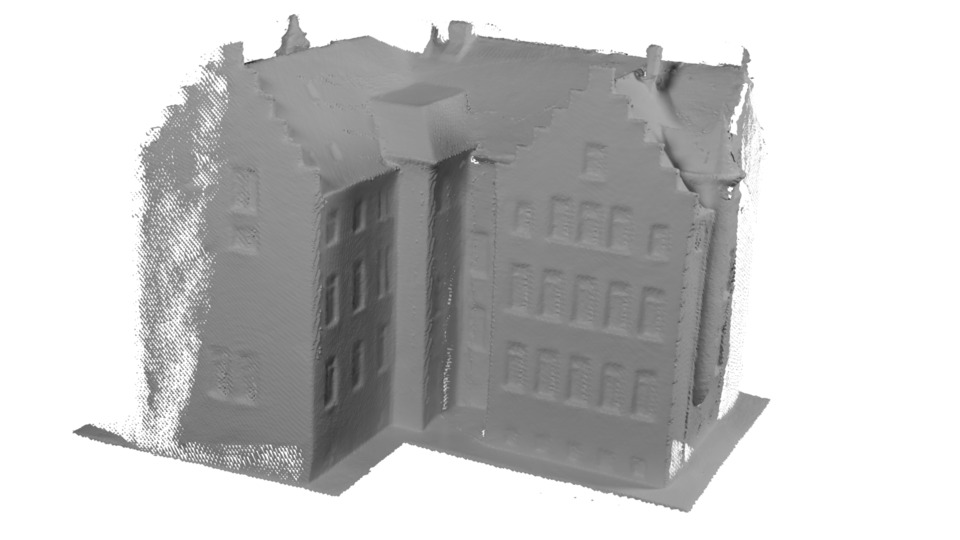}
        \vfill
        {\scriptsize MAtCha Gaussians from 3 views} \\
    \end{minipage}
    \begin{minipage}[b]{0.352\textwidth}
        \centering
        \includegraphics[trim={0 0 0 5.5cm},clip,width=0.98\linewidth]{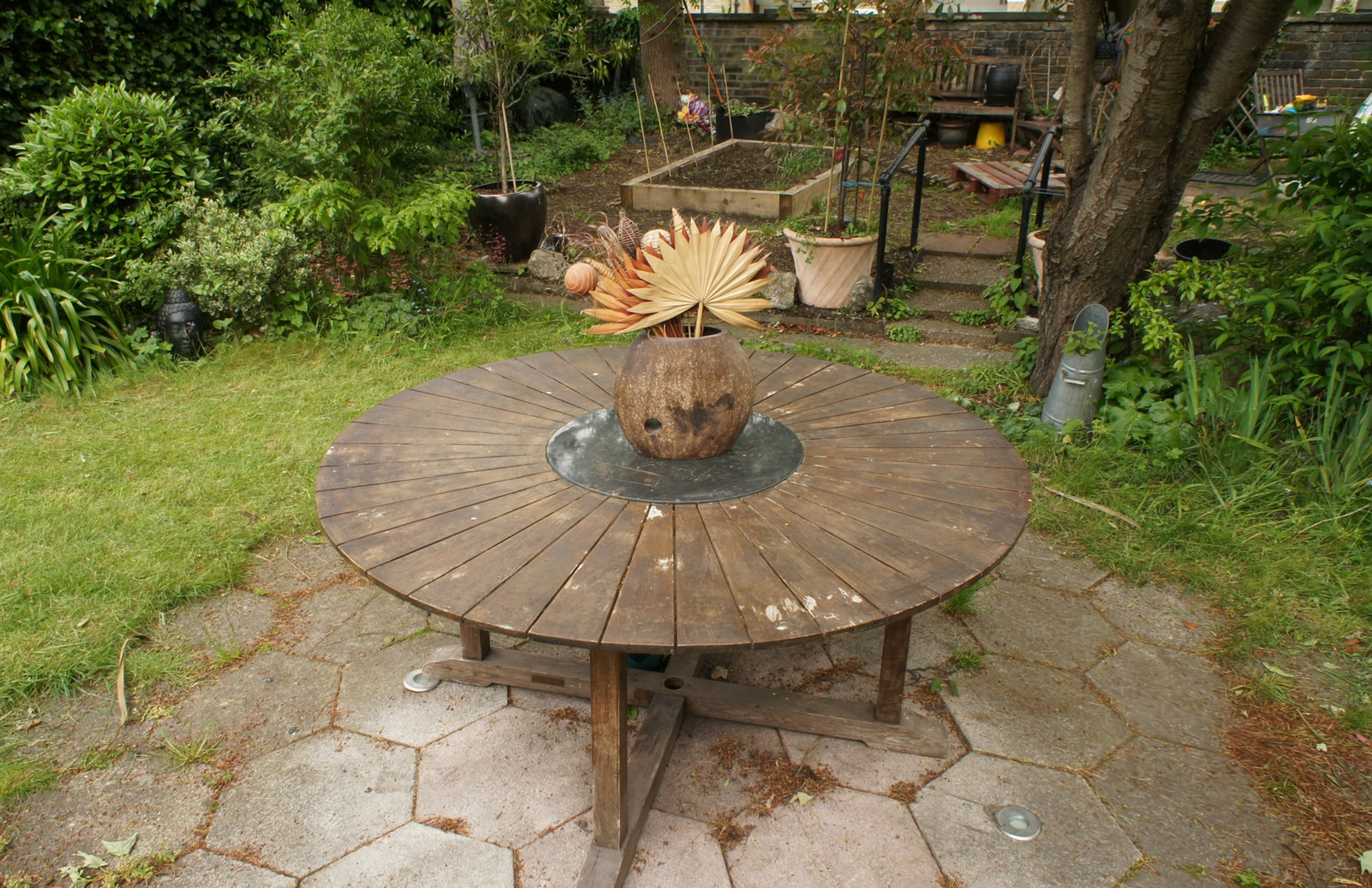}
        \vfill
        \includegraphics[trim={0 1cm 0 2.5cm},clip,width=0.98\linewidth]{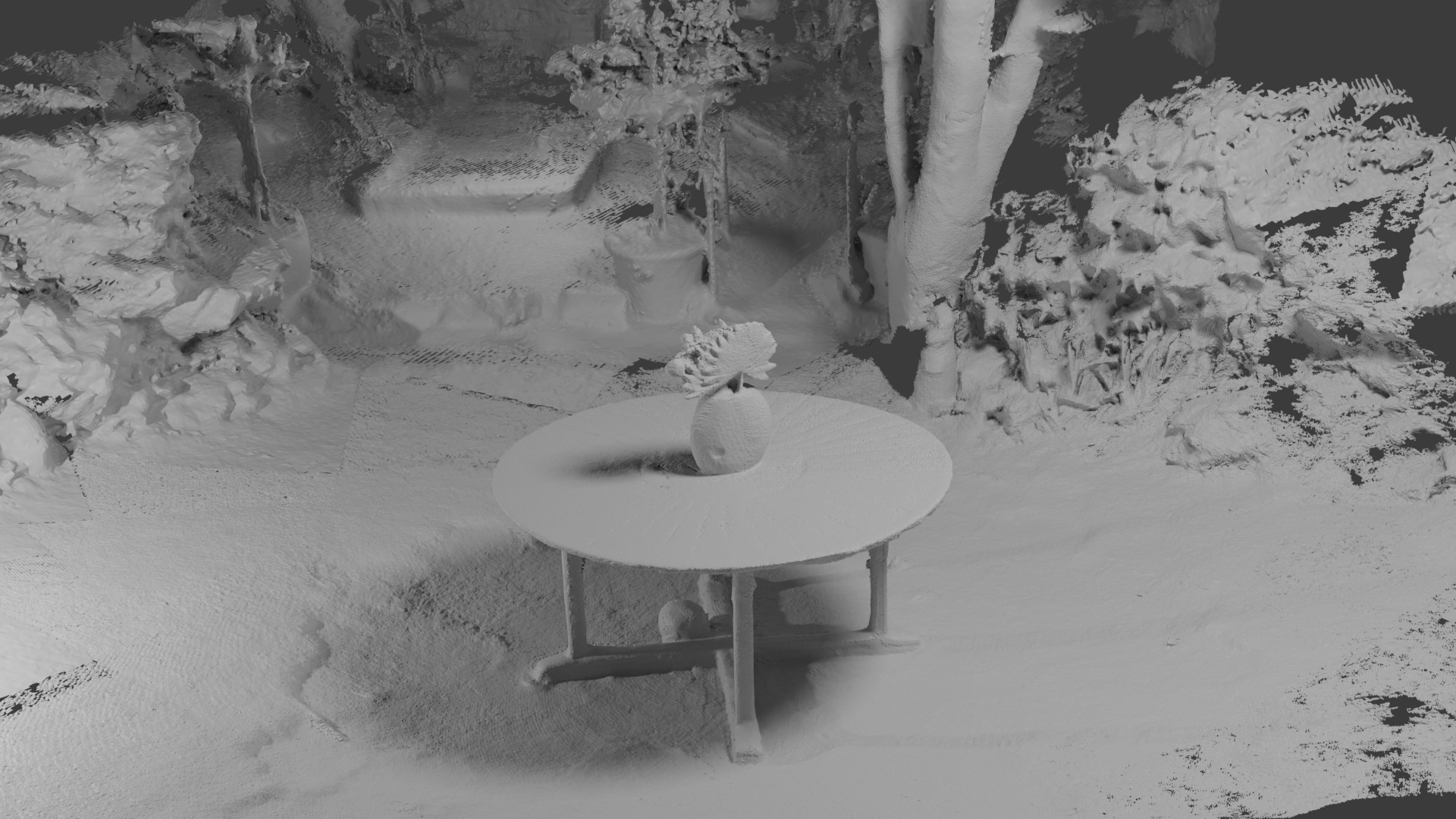}
        \vfill
        {\scriptsize MAtCha Gaussians from 10 views} \\
    \end{minipage}
    \begin{minipage}[b]{0.32\textwidth}
        \centering
        \includegraphics[width=0.98\linewidth, height=0.55\linewidth]{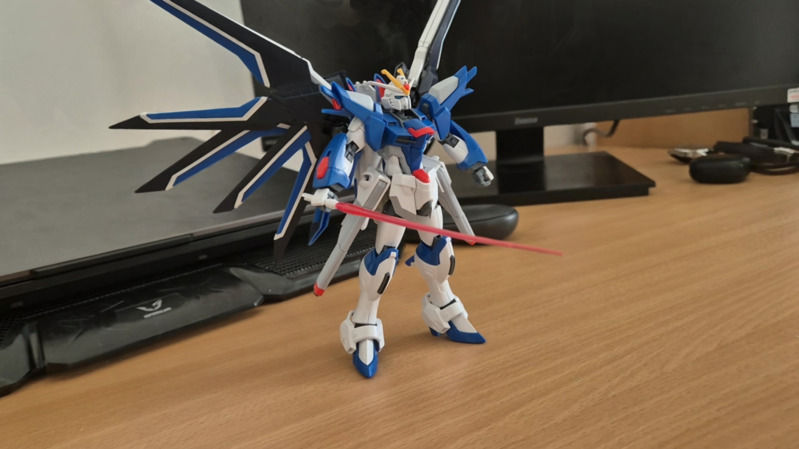}
        \vfill
        \includegraphics[width=0.98\linewidth, height=0.55\linewidth]{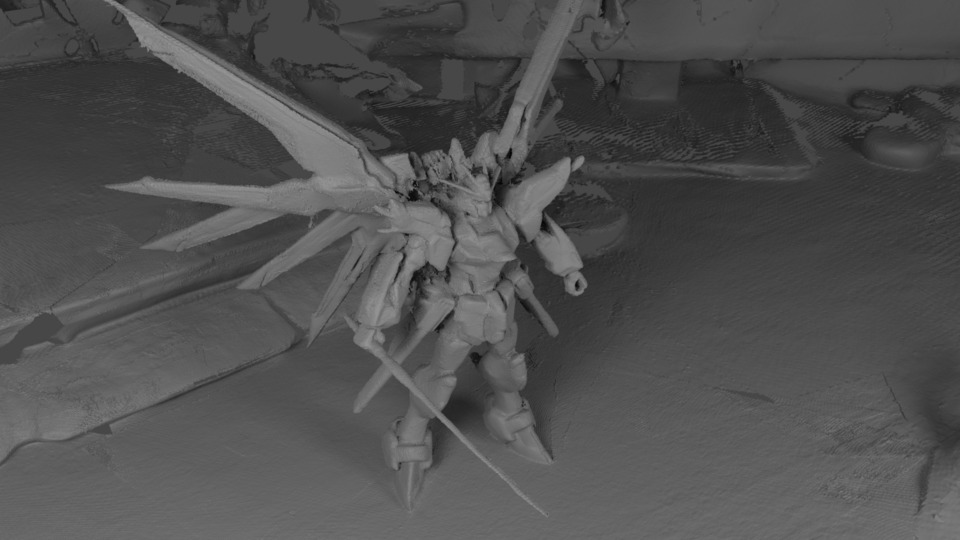}
        \vfill
        {\scriptsize MAtCha Gaussians from 10 views}
    \end{minipage}

    \captionof{figure}{
    \textbf{We propose \method, a novel surface representation for reconstructing high-quality 3D meshes with photorealistic rendering from sparse-view images.}
    Our key idea is to model the underlying scene geometry as an Atlas of Charts in 2D image planes, which we render with 2D Gaussian surfels. We initialize the charts with a monocular depth estimation model and refine them using differentiable Gaussian rendering and a lightweight neural chart deformation model. Combined with a sparse-view SfM model like MASt3R-SfM~\cite{duisterhof2024mast3rsfm}, MAtCha can recover sharp and accurate surface meshes of both foreground and background objects in unbounded scenes within minutes, from a few unposed RGB images. We used 3 views for training for the left most, and 10 views for the rest.
    }
    \label{fig:opening figure}
\end{center}
}]
\begin{abstract}
We present a novel appearance model that simultaneously realizes explicit high-quality 3D surface mesh recovery and photorealistic novel view synthesis from sparse view samples. 
Our key idea is to model the underlying scene geometry \underline{M}esh as an \underline{At}las of \underline{Cha}rts which we render with 2D \underline{Gaussian s}urfels (MAtCha Gaussians).
MAtCha distills high-frequency scene surface details from an off-the-shelf monocular depth estimator and refines it through Gaussian surfel rendering. The Gaussian surfels are attached to the charts on the fly, satisfying photorealism of neural volumetric rendering and crisp geometry of a mesh model, \ie, two seemingly contradicting goals in a single model. 
At the core of MAtCha lies a novel neural deformation model and a structure loss that preserve the fine surface details distilled from learned monocular depths while addressing their fundamental scale ambiguities. 
Results of extensive experimental validation demonstrate MAtCha's state-of-the-art quality of surface reconstruction and photorealism on-par with top contenders but with dramatic reduction in the number of input views and computational time. We believe MAtCha will serve as a foundational tool for any visual application in vision, graphics, and robotics that require explicit geometry in addition to photorealism. 
\end{abstract}
\section{Introduction}
\label{sec:intro}

The neural revitalization of volume rendering has revolutionalized novel view synthesis of real-world scenes. Neural Radiance Fields (NeRF) \cite{mildenhall2020nerf}, 3D Gaussian Splatting \cite{kerbl3Dgaussians} and their many variants provide reliable means to extract appearance representations of intricate scenes that used to be out of our reach, \eg, a flower bouquet. The volume rendering formulation is essential for this as it offers reliable gradients for end-to-end neural function fitting. 

The learned NeRF or Gaussians are, however, fundamentally trained for photorealism in the end results, \ie, 2D images. 
The physical scene representation that one can tease out from these neural representations are only crude approximations of reality. Most notable, the geometry is blurry as it is accurate only up to what is necessary for volume rendering. For instance, it is easy for the networks to distribute the view-dependent reflectance of a surface with multiple 3D Gaussians of unique colors scattered around the surface. The same applies to NeRF, too.

Explicit extraction of the scene geometry would be of natural importance for many applications in vision, robotics, and graphics, as interaction with the scene including editing is often the end goal. 
Extraction of explicit geometry from learned neural representations has typically been achieved by applying algorithms such as TSDF~\cite{curless1996volumetric}, marching cubes~\cite{lorensen1987marching} or screened Poisson surface reconstruction~\cite{kazhdan13screenedpoisson} to the learned volumetric representation. This, however, fundamentally forces a sequential process, in which a volumetric appearance model first needs to be learned. This is suboptimal in two fundamental ways.

The first is that it likely requires view samples (\ie, input images) more than necessary. The dense view sampling for NeRF and Gaussian Splatting are essential to learn volumetric representations for photorealistic view synthesis, but for geometry reconstruction our rich literature in computer vision suggests that much less should suffice.
The second is that, even after this sequential process, we cannot recover accurate scene geometry, especially pertaining to its high-frequency changes, \eg, sharp corners and edges, as it is extracted from an already low-pass filtered representation due to the volume rendering formulation.

In this paper, we ask, can we learn a neural scene representation that realizes photorealistic rendering, on par with leading volumetric representations, but at the same time enables conversion to accurate and sharp scene geometry? Better yet, can we achieve this within minutes and from a much smaller number of images? We answer these fundamental questions in the affirmative with a novel 
neural appearance model  which we refer to as \textbf{MAtCha~Gaussians} or \textbf{MAtCha} for short. 
MAtCha stands for \textit{\textbf{M}esh as an \textbf{At}las of \textbf{Cha}rts}. It models the surface as a 2D manifold in 3D space, \ie, an \emph{atlas} of $n$ \textit{charts}.

\shortmethod offer three key benefits as a learnable scene representation. First, it can be initialized with a monocular depth estimator. This enables distillation of high-frequency surface details from a pre-trained model.  
Second, it lets us perform surface refinement in 2D rather than 3D, which can be achieved efficiently with a lightweight neural deformation model. This enables fast and stable scene optimization from just a few images. 
Finally, it serves as a basis for photorealistic rendering with 2D Gaussian surfels aligned with the charts on the fly. Gaussian surfel rendering provides better gradients compared with mesh rendering for geometry refinement as well as novel view synthesis even with a limited number of input views.

\shortmethod is initialized with depth estimates from learned monocular models. Monocular depth estimators, however, suffer from scale ambiguities. Scale variation across views lead to erroneous 3D scene structure, although its derivatives are unaffected. We address this depth ambiguity by introducing a neural deformation model that deforms view-dependent depth estimates to match and align while preserving the high-frequency details. We achieve this with a tiny MLP that takes in a feature vector sampled from a sparse 2D grid in the image space (charts encodings) and a depth dependent feature (depth encodings). The sparsity of the 2D grid ensures that the MLP deforms only low-frequency scene structure for regions with similar depths.

Given a sparse set of RGB images and charts initialized by a monodepth model, we first optimize the neural deformation model for each chart using surface points recovered with structure-from-motion (SfM). These charts are then refined by differentiable Gaussian rendering and a photometric loss. To further preserve the high-frequency information in the monodepth estimates, we impose a structure loss that encourages the deformed charts to maintain the normals and curvatures computed from the derivatives of the initial depth. After this refinement, a unified surface mesh of both foreground and background of the scene can be recovered from our geometrically accurate charts using our two custom methods, multi-resolution TSDF fusion and adaptive tetrahedralization.

We validate the effectiveness of \shortmethod with an extensive set of experiments. We demonstrate that it can recover accurate scene geometry from sparse RGB images within minutes. This is in contrast to current state-of-the-art methods that require dense view sampling and long training time.
We also show that \method can render high-quality images from novel viewpoints in the sparse-view scenarios where existing sparse-view Gaussian Splatting methods suffer from little overlap of views.
Ablation studies show that our proposed deformation model is crucial for accurate surface reconstruction.

We believe \method seamlessly integrates our rich history of 3D geometry reconstruction research into cutting-edge neural representations for appearance modeling and can serve as a foundational tool for a wide range of downstream application domains in vision, graphics, robotics, and beyond.

\section{Related Work}
\label{sec:related}

\newcolumntype{C}{>{\centering\arraybackslash}X}
\begin{table*}[t]
    \centering
    \scriptsize
    \begin{tabularx}{\linewidth}{cCCCCC}
        \hline
        Method & Sparse view & Surface reconstruction & Explicit surface optimization & Reconstruction of unbounded scenes & Fast training ($<$15min) \\ \hline
        Gaussian Surfel~\cite{dai2024high}, 2DGS~\cite{huang20242d}, GOF~\cite{yu2024gaussian} & & \checkmark &  & \checkmark \\
        SuGaR~\cite{guedon2024sugar}, Gaussian Frosting~\cite{guedon2024gaussian} & & \checkmark & & \checkmark & \\
        InstantSplat~\cite{fan2024instantsplat} & \checkmark & & & \checkmark & \checkmark \\
        SparseNeus~\cite{long22sparseneus}, VolRecon~\cite{ren23volrecon} & \checkmark & \checkmark & & & \checkmark \\
        S-VolSDF~\cite{wu23svolsdf}, NeuSurf~\cite{huang24neusurf}, Spurfies~\cite{raj2024spurfies} & \checkmark & \checkmark & & & \\
        \textbf{Ours (\method)} & \checkmark & \checkmark & \checkmark & \checkmark & \checkmark  \\ \hline
    \end{tabularx}
    \caption{\textbf{Comparisons between our method and existing methods for image-based 3D reconstruction.} Our method achieves fast reconstruction of unbounded scene mesh from sparse-view images by directly optimizing explicit surface manifolds.
    \koheirmk{We may be able to remove dense-view methods from this, and instead focus on comparison with sparse methods.}\koheirmk{Add something like "scenes with background objects"?}\ichikawarmk{Should we keep "Mesh equipped with UV maps? UV map of charts enables us to instantiate Gaussians but we cannot use it for downstream applications for now.}}
    \label{tab:methods comparison}
\end{table*}

\cref{tab:methods comparison} summarizes the key contributions of our method in comparison with previous novel view synthesis and surface reconstruction methods.

\vspace{-8pt}
\paragraph{Structure from Motion}
Structure-from-Motion (SfM)~\cite{LONGUET-HIGGINS-1987-sfm-seminal,snavely-2006-structure-from-motion,schonberger2016structure} estimates extrinsic and intrinsic camera parameters and reconstructs a sparse 3D point cloud from uncalibrated multi-view images.
Recent differentiable SfM methods~\cite{wang2024dust3r,wang2024vggsfm,brachmann2024scene, leroy2024grounding,duisterhof2024mast3rsfm} demonstrate impressive results on complex real-world scenes.
SfM-reconstructed point clouds are, however, sparse and do not have sufficient high-frequency details as the underlying formulation fundamentally relies on multi-view correspondences. This is fatal for photorealistic rendering and accurate surface reconstruction. SfM can instead provide strong initializations for these tasks.

\vspace{-8pt}
\paragraph{Novel View Synthesis}
NeRF~\cite{mildenhall2020nerf}, 3D~Gaussian~Splatting~\cite{kerbl3Dgaussians}, and their derivatives~\cite{mueller2022instantngp,yu2021plenoxels,chen2022tensorf,barron2023zipnerf,barron2021mip,yu2024mip} achieve impressive photorealism in novel view synthesis. 
They optimize a representation based on a neural implicit function or a set of Gaussian primitives with differentiable volume rendering.
These methods require very dense view samples to learn an accurate scene representation. Recent works have introduced various approaches for learning these representations from sparser views, for instance by applying regularizations on 3D Gaussians~\cite{yu2024lm,zhang2024cor,li2024dngaussian,paliwal2024coherentgs,zhu2024fsgs,jiang2024construct,xwz2024eccv,fan2024instantsplat} or by training a feed-forward deep network to directly estimate 3D Gaussian parameters~\cite{charatan2024pixelsplat,chen2024mvsplat,xu2024depthsplat}. 
These methods, however, are focused on novel view synthesis and surfaces extracted from them remain inaccurate and noisy.

\vspace{-8pt}
\paragraph{Surface Reconstruction from RGB images}
Recent image-based surface reconstruction methods also leverage differentiable volume rendering~\cite{wang2021neus,guedon2024sugar,guedon2024gaussian,dai2024high,huang20242d,yu2024gaussian,zhang2024rade,chen2024pgsr,chen2023neusg,zhang2024neural,lyu20243dgsr,yu2024gsdf,turkulainen2024dn,chen2024vcr,wolf2024gsmesh}. 
For instance, Gaussian~Surfels~\cite{dai2024high} and 2D~Gaussian~Splatting~(2DGS)~\cite{huang20242d} use flat 2D~Gaussians instead of 3D~Gaussians to represent the surface accurately. 
These methods, however, require dense view sampling to constrain the millions of tiny Gaussians. 

A few methods handle sparse inputs. SparseNeus~\cite{long22sparseneus}, VolRecon~\cite{ren23volrecon}, and S-VolSDF~\cite{wu23svolsdf} exploit pretrained feed forward networks for multi-view inputs. 
NeuSurf~\cite{huang24neusurf} first reconstructs global structures from an SfM point cloud, then refines local geometry by fitting a signed distance function~(SDF) with local feature consistency.
Spurfies~\cite{raj2024spurfies} leverages local priors from a pretrained geometry decoder for optimizing an SDF.

Multi-view feed-forward networks and geometry decoders, however, struggle to generalize to unseen, unbounded scenes, as they are trained with a limited number of object-centric or synthetic datasets. 
Additionally, the optimization of a signed distance field might cause loss of geometry details due to ineffective constraints on the 3D volumetric representation.
To the best of our knowledge, no existing method can reconstruct sharp meshes of unbounded scenes from sparse input views.

\method fills this hole. Its explicit surface representation exploits and preserves local geometry details obtained from a monocular depth estimation model for fast and sharp surface reconstruction from sparse-view images.

\section{Preliminaries}

Let us first recall some preliminaries for surface representations.

\vspace{-8pt}
\paragraph{Gaussian Splatting}
Gaussian~Splatting~\cite{kerbl3Dgaussians} and its derivatives model 3D scenes with large collections of tiny, smooth 3D ellipsoids.
They realize efficient rasterization-based volume rendering which enables robust learning from RGB images. 
More recently, their 2D variants, namely 2D Gaussian surfel representations have been proposed~\cite{huang20242d,dai2024high}.
All these representations obviously have an extremely large degree of freedom. Each of the numerous Gaussians is free to slightly move away from the true surface to approximate view-dependent appearance. This fundamental redundancy causes even the most recent methods to struggle to recover accurate surfaces from sparse-view images.

\vspace{-8pt}
\paragraph{Surfaces as 2D Manifolds}
Let $M$ be a subset of $\IR^3$, typically a surface. A \emph{chart} $(U, \phi)$ on $M$ consists of an open subset $U \subset M$ equipped with a homeomorphism $\phi:U \rightarrow \IR^2$. In other words, a chart is a continuous bijection between $U$ and an open subset of $\IR^2$, implying that the subset $U$ of $M$ can be represented by continuously deforming a flat surface patch.

The set $M$ is called a \emph{2-dimensional manifold} in $\IR^3$, if there exists a collection of $n$ charts $(U_i, \phi_i)_{0 \leq i < n}$ covering $M$, \ie, such that $\bigcup_i U_i = M$. The collection of charts $(U_i, \phi_i)_i$ is called an \emph{atlas} on $M$. 
Intuitively, a 2-dimensional manifold is a subset of $\IR^3$ that can be represented by deforming and patching together a collection of planar pieces.
In this paper, we will denote charts by $\phi_i:U_i \rightarrow V_i$ where $V_i \subset \IR^2$, and their inverse mapping by $\psi_i:=\phi_i^{-1}:V_i \rightarrow U_i$.

A \emph{UV map} is an example of a chart which is widely-used for representing the texture of a surface. 
Given a texture image, a UV map maps every vertex of the mesh to a single point in the image, allowing for texturing the 3D surface by looking up the 2D texture.

\section{\method}

Let us derive \textbf{MAtCha~Gaussians}, an appearance model learnable from a sparse set of $N$ RGB images, from which a detailed surface mesh can be extracted. MAtCha is a 2D manifold equipped with Gaussians. 
Specifically, we model the surface of the scene as a collection of $n\leq N$ charts, each chart corresponding to one of the input views. 
There are three key benefits of using this representation, particularly with sparse views.

First, we can directly initialize the charts by using detailed depth maps computed with a pre-trained monocular depth estimation model (monodepth model)~\cite{Yang24depthanything,depth_anything_v2} and explicitly distill the high-frequency geometry captured by the depth maps into our representation. 
Second, it allows us to optimize the surface with 2D deformation maps instead of dense 3D grids, resulting in a significantly more efficient optimization. It is also more robust, as we can leverage a lightweight neural deformation model on the 2D map to efficiently constrain the geometry for sparse-view surface reconstruction. 
Finally, we can refine our explicit surface representation with differentiable volumetric Gaussian rendering by instantiating 2D Gaussian surfels aligned with the charts on the fly, which enables efficient refinement of the manifold and photorealistic rendering. 
Reciprocally, our charts explicitly constrain Gaussians and prevent them from diverging even with very sparse view samples.

\begin{figure*}[t]
    \centering
    \begin{tikzpicture}[
        node distance=0.75cm,  
        arrow/.style={
            ->,
            thick,
            >=stealth,
            color=gray!90
        },
        detail/.style={
            rectangle,
            rounded corners=3pt,
            minimum height=3cm,
            align=center,
            font=\small,
            inner sep=4pt
        }
    ]
    
        \definecolor{chartcolor}{RGB}{255,240,230}  
        \definecolor{gaussiancolor}{RGB}{230,255,235}  
        \definecolor{mlpcolor}{RGB}{240,170,170}  
        \definecolor{imagebox}{RGB}{230,240,255}  
        \definecolor{pluscolor}{RGB}{190,190,240}  
        \definecolor{geometry}{RGB}{230,255,235}
        \definecolor{chartcolor_dark}{RGB}{191,180,172}  
        
        \node[anchor=center] (input) {
            \begin{tikzpicture}
                \node[anchor=center] at (-0.15,0.3) {\includegraphics[width=1.8cm]{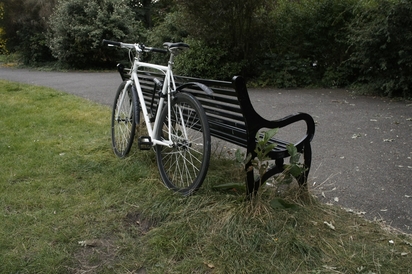}};  
                \node[anchor=center] at (0,0.15) {\includegraphics[width=1.8cm]{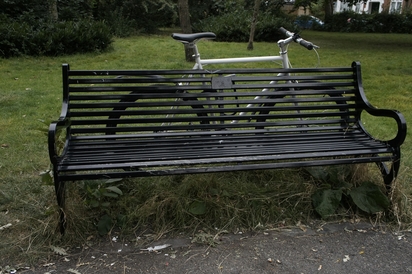}};
                \node[anchor=center] at (0.15,0) {\includegraphics[width=1.8cm]{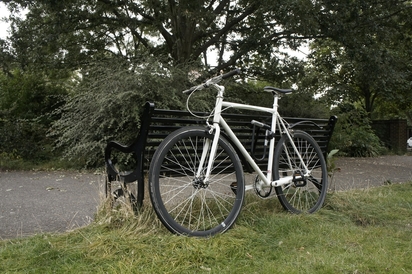}};
            \end{tikzpicture}
        };

        \node[anchor=center, right=0.8cm of input] (depth) {
            \begin{tikzpicture}
                \node[anchor=center] at (-0.15,0.3) {\includegraphics[width=1.8cm]{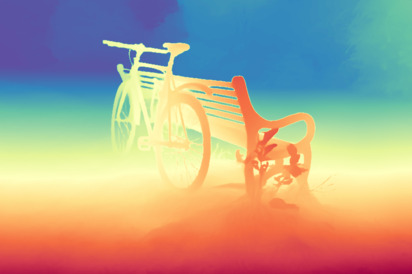}};
                \node[anchor=center] at (0,0.15) {\includegraphics[width=1.8cm]{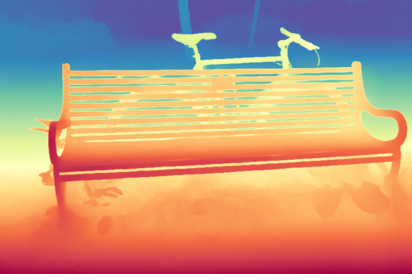}};
                \node[anchor=center] at (0.15,0) {\includegraphics[width=1.8cm]{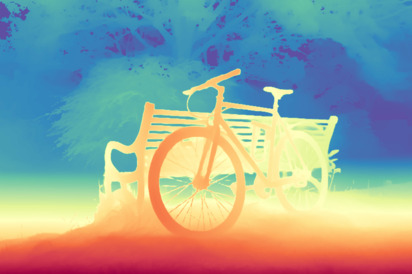}};
            \end{tikzpicture}
            \hspace{0.05cm}  
            \includegraphics[width=2.2cm]{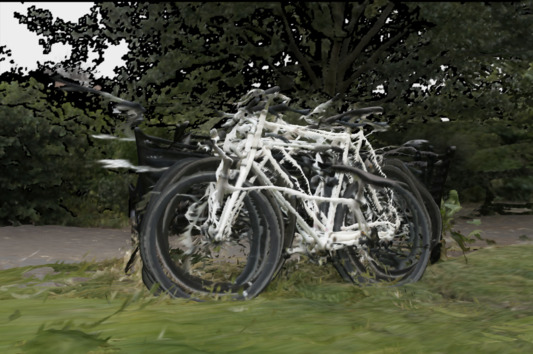}
        };
        
        \node[anchor=center, right=0.8cm of depth] (aligned) {
            \includegraphics[width=2.2cm]{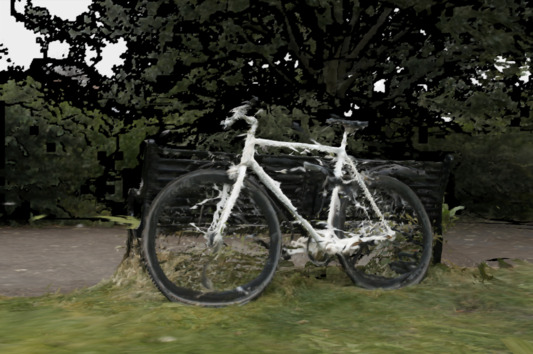}  
        };
        
        \node[anchor=center, right=0.8cm of aligned] (final) {
            \includegraphics[width=2.2cm]{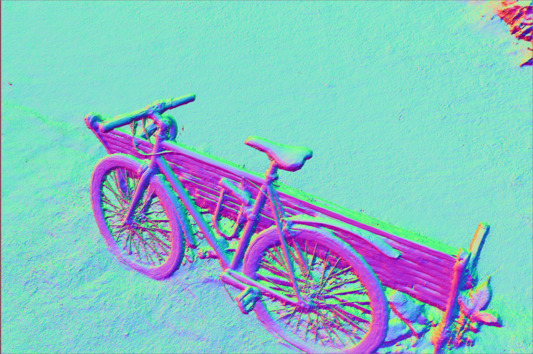}
            \hspace{0.05cm}  
            \includegraphics[width=2.2cm]{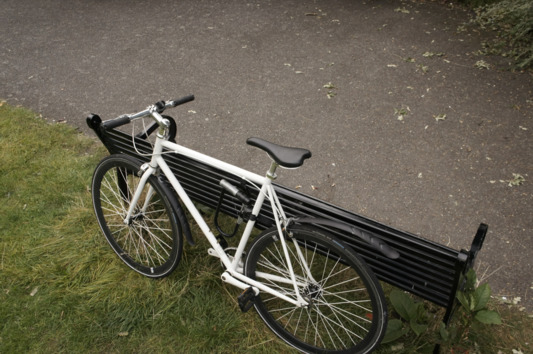}
        };
        
        \node[below=-0.1cm of input, text width=1.8cm, align=center, font=\footnotesize] {Sparse RGB Images};
        \node[below=-0.1cm of depth, text width=3.8cm, align=center, font=\footnotesize] {Depth Maps and Initial Charts};
        \node[below=-0.01cm of aligned, text width=2.2cm, align=center, font=\footnotesize] {Aligned Charts};
        \node[below=-0.01cm of final, text width=3.8cm, align=center, font=\footnotesize] {Refined Charts equipped with Gaussians};
        
        \draw[arrow, very thick] (input.east) -- (depth.west) node[midway,above,font=\scriptsize] {Monodepth};
        \draw[arrow, very thick] (depth.east) -- (aligned.west) node[midway,above,font=\scriptsize] {SfM data};
        \draw[arrow, very thick] (aligned.east) -- (final.west) node[midway,above,font=\scriptsize] {Images};
        
        \path (depth.east) -- (aligned.west) coordinate[midway] (arrow1);
        \path (aligned.east) -- (final.west) coordinate[midway] (arrow2);
        
        \node[detail, fill=chartcolor, below=2cm of arrow1, xshift=-4cm,
         text width=9cm, 
        ] (detail1) {
            \textbf{Optimizing Charts with a Robust Deformation Model}
            
            \begin{tikzpicture}[
                mlpbox/.style={draw, rectangle, minimum width=2cm, minimum height=1.3cm, font=\small, fill=mlpcolor},  
                imgbox/.style={draw, rectangle, minimum width=1.3cm, minimum height=1.3cm, font=\scriptsize, fill=imagebox},  
                arrow2/.style={->, >=stealth, thick, color=black}
            ]
                \node[imgbox, inner sep=2pt] (grid) at (0,0.5) {
                    \begin{tabular}{c}
                        \includegraphics[width=0.8cm]{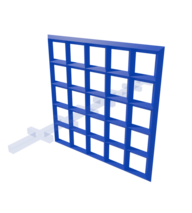} \\
                        Charts encodings
                    \end{tabular}
                };
                
                \node[imgbox, inner sep=2pt] (depth_enc) at (0,-1.) {
                    \begin{tabular}{c}
                        \includegraphics[width=0.8cm]{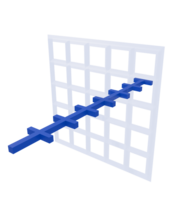} \\
                        Depth encodings
                    \end{tabular}
                };
                
                \node[circle,draw,inner sep=0.05cm,minimum size=0.2cm,fill=pluscolor] (plus1) at (1.55,-0.6) {$+$};
                
                \node[mlpbox] (mlp) at (3.2,0.5) {MLP};
                
                \node[imgbox, fill=geometry, inner sep=2pt] (image1) at (3.2,-1) {
                    \begin{tabular}{c}
                        \includegraphics[width=1.4cm]{images/pipeline/initial_charts.jpg} \\
                        Initial Charts
                    \end{tabular}
                };
                
                \node[circle,draw,inner sep=0.05cm,minimum size=0.2cm,fill=pluscolor] (plus2) at (4.5, -0.6) {$+$};
                
                \node[imgbox, fill=geometry,inner sep=2pt] (finalimage) at (6.3,-0.12) {
                    \begin{tabular}{c}
                        \includegraphics[width=1.4cm]{images/pipeline/aligned_rgb.jpg} \\
                        Deformed Charts
                    \end{tabular}
                };
                
                \draw[arrow2] (grid.east) -| (plus1.north);
                \draw[arrow2] (depth_enc.east) -| (plus1.south);
                \draw[arrow2] (plus1.east) -- ++(0.15,0) |- (mlp.west);
                \draw[arrow2] (mlp.east) -| (plus2.north);
                \draw[arrow2] (image1.east) -| (plus2.south);
                \draw[arrow2] (plus2.east) -- (finalimage.west);
            \end{tikzpicture}
            
        };
        
        \node[detail, fill=gaussiancolor, below=2cm of arrow2, xshift=1.5cm, text width=7cm, minimum height=3.5cm] (detail2) {
            \textbf{Rendering Charts with 2D Gaussian Splatting} \\
            \begin{tabular}{c}
                \includegraphics[height=2.07cm]{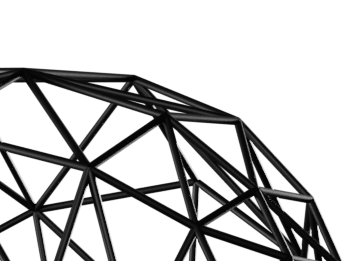}
                \includegraphics[height=2.07cm]{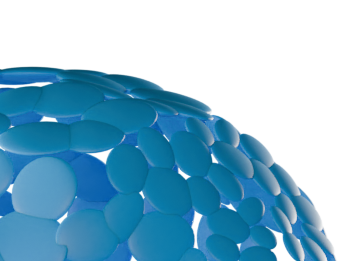} \\
                We combine our deformation model with Gaussians\\
                instantiated on the surface to refine the geometry. 
            \end{tabular}
        };
        
        \draw[gray!90, dashed, very thick]  (detail1.north) |- ($(arrow1)+(0,-1.7)$) -- (arrow1);
        \draw[gray!90, dashed, very thick] (detail1.north) |- ($(arrow2)+(-0.2,-1.7)$) -- ($(arrow2)+(-0.2,0)$) ;
        \draw[gray!90, dashed, very thick] ($(arrow2)+(0.1,0)$) -- ($(arrow2)+(0.1,-1.7)$) -| (detail2.north);
    \end{tikzpicture}
    \caption{
    \textbf{Overview of \method.} Given a few RGB images and their camera poses obtained using a sparse-view SfM method such as MASt3R-SfM~\cite{duisterhof2024mast3rsfm}, we first initialize charts using a pretrained monocular depth estimation model. Each chart is represented as a mesh equipped with a UV map, mapping a 2D plane to the 3D surface. 
    We then optimize our charts and enforce their alignment with input SfM data using two key components: (1) 1D depth encodings for quickly aligning the initial depth maps together, and (2) charts encodings for efficiently deforming the geometry while preserving surface details. 
    Our aligned charts provide a sharp, dense and accurate estimate of the 3D scene, which can be further refined using input images and a Gaussian Splatting-based rendering pipeline. Our representation allows for reconstructing high-quality surface meshes within minutes, even in sparse-view scenarios.
    }
    \label{fig:pipeline}
\end{figure*}

\cref{fig:pipeline} depicts the overall pipeline of \method. Given a sparse set of $N$ RGB images $I_i$ and their corresponding cameras $c_i$ with extrinsics $(R_i, t_i) \in SO(3) \times \IR^3$ and intrinsics $K_i \in \IR^{3\times 3}$, we optimize a set of charts so that it approximates the true geometry of the scene. Note that, in practice, we can obtain the camera parameters even for sparse unposed images using a sparse structure-from-motion (SfM) method~\cite{duisterhof2024mast3rsfm}. We first initialize the charts using a monodepth model. Then, we align the charts with surface points recovered by the SfM method using our novel deformation model. Finally, we refine the charts with differentiable rendering based on Gaussian surfels.

\subsection{Chart Initialization with MonoDepth Estimates}
For a given number of charts $n \leq N$, we initialize the charts by using depth maps $(D_i)_{0\leq i<n}$ estimated with a pre-trained monodepth model~\cite{depth_anything_v2} from the input views $I_i$.
A depth map is indeed a mapping from a 2D plane to 3D scene points, making it a natural candidate for representing a chart. 
In practice, in sparse view scenarios, we set $n=N$ and backproject all depth maps to 3D space.

We denote the initial chart constructed by backprojecting the depth map $D_i$ to 3D space with $\psi^{(0)}_i: V_i \rightarrow U^{(0)}_i$. 
The mapping $\psi^{(0)}_i$ maps the 2D UV coordinates in $V_i\subset [0,1]^2$ to some 3D points in $U_i^{(0)}\subset \IR^3$, where $U_i^{(0)}$ is supposed to be a subset of the true surface of the scene.

The relative scales of these initial charts computed from monodepth results are, however, generally inaccurate. Simply backprojecting depth maps into 3D space results in a chaotic, unaligned manifold and does not provide an accurate enough initial estimate of the true surface.
We can try to estimate the relative scales between the initialized charts and the true surface, by exploiting surface points recovered by the SfM method. 
Previous works~\cite{turkulainen2024dn,kerbl2024hierarchicalgaussians} model the discrepancies with simple models such as a global affine rescaling of the depth maps. Although easy to compute, such an approach inevitably results in poor accuracy due to differences in the relative scales between objects.
On the other hand, pixel-wise scaling does not work in sparse view scenarios as its over-parameterization will cause loss in high-frequency of the geometry.

\subsection{Lightweight Chart Deformation Model}

To refine and align the initialized charts by resolving the complex object-dependent scaling while preserving the high-frequency information from the monodepth estimates, we introduce a novel deformation model based on what we refer to as \emph{chart encodings}. 

\vspace{-8pt}
\paragraph{Chart Encoding} For deforming each chart $i$, we maintain 1) a sparse 2D grid of learnable features $E_i \in \mathbb{R}^{r h \times r w \times d}$ in UV space, where $r$ is a size ratio, $h$ and $w$ are the height and width of the depth map, and $d$ is the feature dimension; and 2) a tiny MLP $f_{\theta_i}: \mathbb{R}^d \rightarrow \mathbb{R}^3$ that maps these features to 3D deformation vectors.

The deformation field for chart $i$ at UV coordinate $u$ is 
\begin{equation}
    \Delta_i(u) = f_{\theta_i}\left[E_i(u)\right]\,,
\end{equation}
where $E_i(u)$ bilinearly interpolates features from the sparse grid at coordinate $u$. The deformed inverse map $\psi_i$ which maps the UV coordinates in $V_i$ to 3D points on the updated surface $U_i$ becomes
\begin{equation}
    \psi_i(u) = \psi_i^{(0)}(u) + \Delta_i(u)\,.
\end{equation}
The sparsity of the 2D feature grid encourages the 2D deformation field to contain only low-frequency deformation, \ie, the high-frequency structures in the initial charts are preserved during the optimization.

\vspace{-8pt}
\paragraph{Depth Discontinuities}
The deformation field, however, needs to be discontinuous at contours of the objects, due to inconsistent scales between objects in the initial charts.

To model such discontinuities,
we augment our chart encodings with an additional depth-dependent feature which we refer to as \emph{depth encodings}. For each UV coordinate $u$, we compute an encoding $z_i(d(u))$ that depends only on the initial depth value of the pixel $d(u) = (P_i \circ \psi_i^{(0)}(u))_z$ (\ie, the z-component of the backprojected point), where $P_i$ is the function transforming 3D points to the coordinate frame of the depth map $D_i$. 
These features are stored along the depth axis and interpolated depending on the depth of the point. The complete feature vector used for deformation becomes
\begin{equation}
    \Delta_i(u) = f_{\theta_i}\left[
    E_i(u) + z_i(d(u)
    )\right]\,,
\end{equation}
for any 2D point $u$ in the UV space. The feature $z_i(d(u))$ acts as positional encoding that allows points at different depths to be deformed independently, even if they are close in the UV space. The depth encoding helps disambiguate spatial relationships that are not captured by the 2D chart encoding alone, leading to more accurate surface reconstruction. 
It also acts as a useful prior by enforcing points with similar depths to be deformed similarly, which is important particularly for sparse view samples. 
Combining features stored in sparse 2D grids and along the depth axis has two main advantages. It requires less memory than storing a full 3D grid of features as it has quadratic space complexity, and simultaneously makes the deformation model more robust to sparsity in the input data.

\subsection{Aligning the Manifold with SfM Points}

With our neural deformation model, we first optimize its weights to make our charts fit as much as possible with the SfM surface points, while maintaining the detailed structure as originally captured by the depth maps. We also encourage the charts to align together to form a coherent and unified manifold. 
We achieve this with the following losses.

\paragraph{Fitting loss $\mathcal{L}_{\text{fit}}$} We encourage the charts to fit the SfM points by minimizing the distance between the SfM points and the deformed charts. Specifically, for each chart, we project the SfM points visible in image $i$ to the UV space of the chart $i$, and we minimize the distance between the SfM points and the corresponding points on the chart: 
\begin{equation}
    \mathcal{L}_{\text{fit}} = \sum_{i=0}^{n-1} \sum_{k=0}^{m_i-1} \|\psi_i(u_{ik}) - p_{ik}\|_1 \,,
\end{equation}
where $u_{ik}$ is the UV coordinate of the $k$-th SfM point visible in image $i$, and $p_{ik}$ is the 3D position of the $k$-th SfM point visible in image $i$. 

In practice, we cannot just rely on a simple fitting loss, as the SfM points may contain outliers.
To address this issue, we introduce for each chart $i$ a learnable confidence map $C_i\in [0,+\infty)^{h\times w}$ that indicates the regions of the chart that are likely to be located on the true surface of the scene.
If the optimization struggles to fit the chart to the SfM points, the confidence map will automatically adjust to downweight the loss in regions where the optimization struggles. Such regions are likely to contain the outlier SfM points.
Specifically, we take inspiration from DUSt3R~\cite{wang2024dust3r} and use a revised fitting loss
\begin{equation}
    \mathcal{L}_{\text{fit}} = \frac{1}{n}
    \sum_{i=0}^{n-1} \sum_{k=0}^{m_i-1} C_i(u_{ik}) \|\psi_i(u_{ik}) - p_{ik}\|_1 -
    \alpha \sum_{i=0}^{n-1} \log(C_i)\,,
\end{equation}
where the second term is a regularization term~\cite{wan18confnet} and $\alpha$ is a hyperparameter. We compute the confidence map as $C_i = 1 + \exp(\hat{C}_i)$, where $\hat{C}_i$ are optimizable parameters.

\paragraph{Structure loss $\mathcal{L}_{\text{struct}}$} We also explicitly encourage the charts to maintain the same sharp structure as the initial depth maps by minimizing the distance between the first and second order derivatives of both the depth maps and the charts. Rather than using all explicit derivatives, we rely on normal and mean curvature regularization, acting respectively on first and second order derivatives
\begin{equation}
    \mathcal{L}_{\text{struct}} = 
    \sum_{i=0}^{n-1} \left( 1 - N_i \cdot N_i^{(0)} \right) + 
    \frac{1}{4}\sum_{i=0}^{n-1} \| M_i - M_i^{(0)} \|_1 \,,
    \label{eq:structure_loss}
\end{equation}
where $N_i$ and $M_i$ are the normal and mean curvatures of chart $i$ computed following~\cite{huang20242d,dai2024high}, and $N_i^{(0)}$ and $M_i^{(0)}$ are those of the initial depth map of chart $i$, respectively.

\paragraph{Mutual alignment loss $\mathcal{L}_{\text{align}}$} We encourage the charts to align together to form a coherent manifold by minimizing the distance between neighboring points located on different charts. Specifically, for each chart $i$, we project the points located on the charts into the screen space of other charts $j$. 
If the corresponding point in $j$ is close enough to the chart $i$, these points are likely to be on the same surface so we minimize the distance between them.
Overall, we compute the mutual alignment loss
\begin{equation}
    \mathcal{L}_{\text{align}} = 
    \sum_{i,j=0}^{n-1} \sum_{u\in V_i} 
    \min\left( \|\psi_i(u) -  \psi_j \circ P_j \circ \psi_i(u)\|_1, \tau \right)\,,
\end{equation}
where $\tau$ is an attraction hyperparameter controlling the maximum distance between points on different charts for considering them to be on the same surface. In contrast to the fitting loss, which may rely only on a sparse set of points depending on the SfM back-end used, the alignment loss acts as a dense regularization on the full surface, helping to form a coherent manifold.

Our complete optimization loss for aligning the charts is
\begin{equation}
    \calL = 
    \calL_{\text{fit}}
    + \lambda_{\text{struct}} \calL_{\text{struct}}
    + \lambda_{\text{align}} \calL_{\text{align}}\,,
\end{equation}
with $\lambda_{\text{struct}}=4$ and $\lambda_{\text{align}}=5$.
This alignment step is very fast and generally takes less than a few minutes to converge.
The charts, however, may still struggle to perfectly align with fine structures. We resolve this by further refining the charts with differentiable rendering.

\begin{figure*}[t]
    \centering
    \begin{minipage}[b]{0.16\textwidth}
        \centering
        \includegraphics[trim={0 1cm 0 4cm},clip,width=0.98\linewidth]{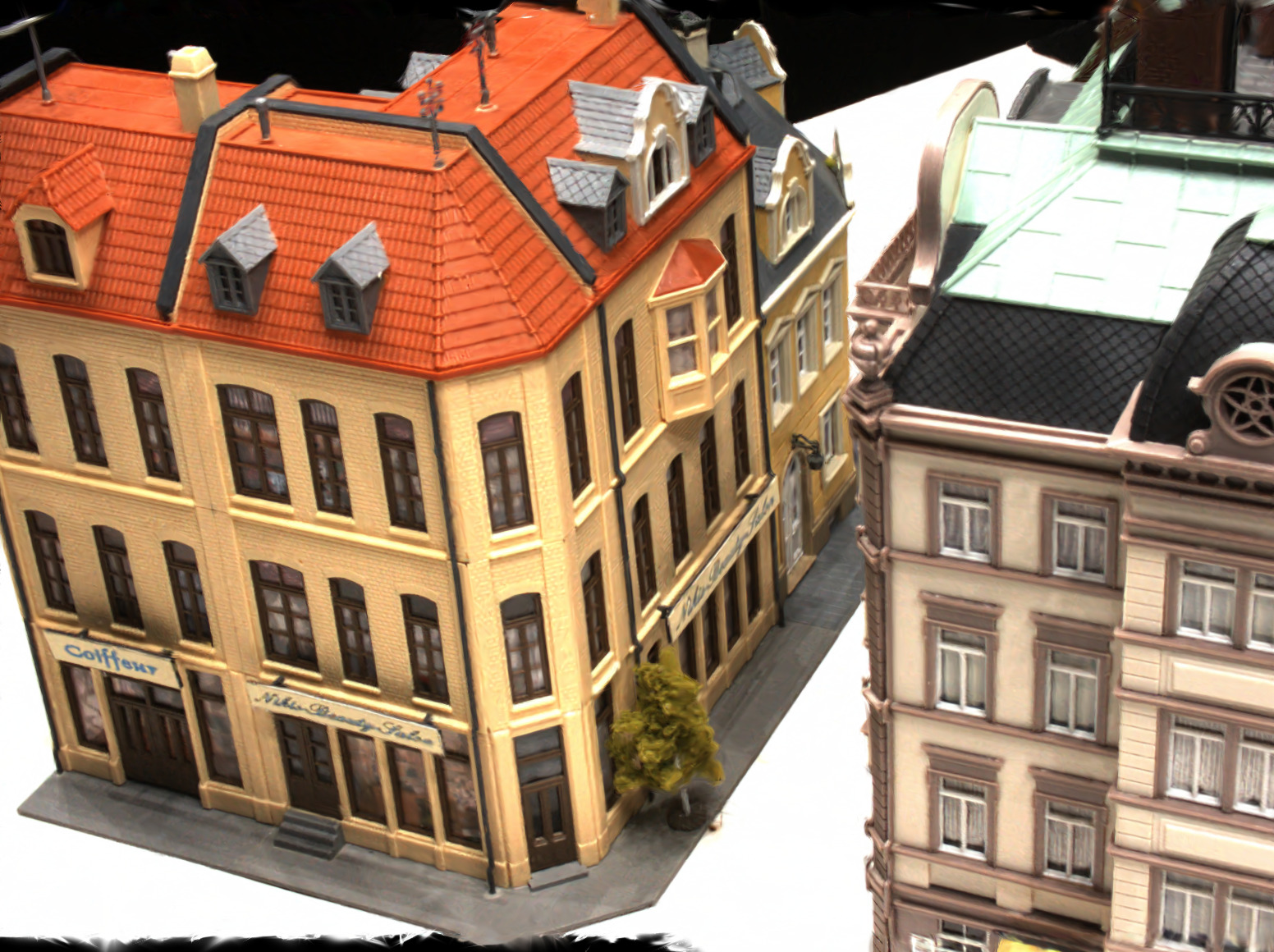}
        \includegraphics[width=0.98\linewidth]{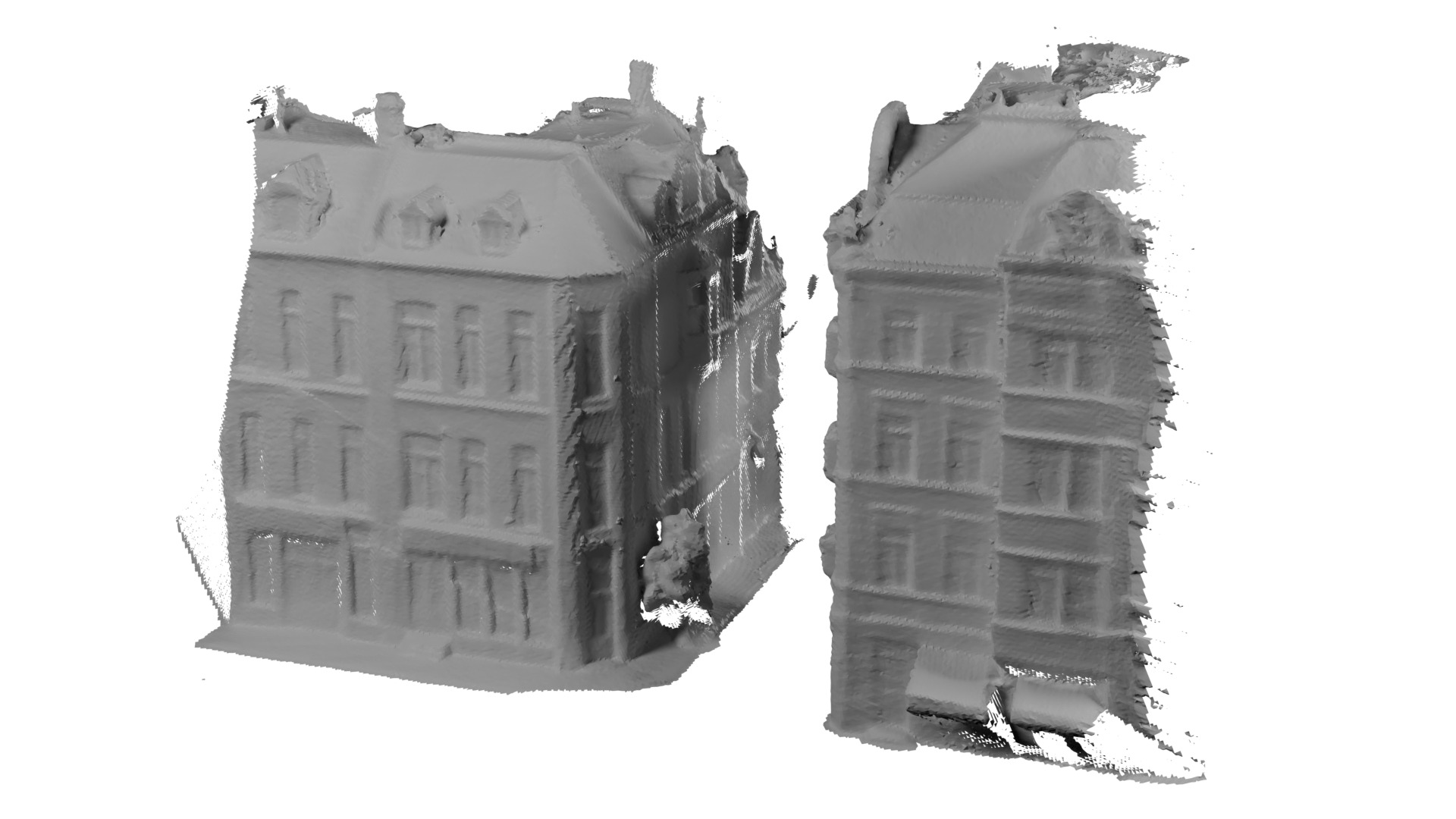}
        \vfill
        3 views
    \end{minipage}
    \begin{minipage}[b]{0.16\textwidth}
        \centering
        \includegraphics[trim={0 1cm 0 4cm},clip,width=0.98\linewidth]{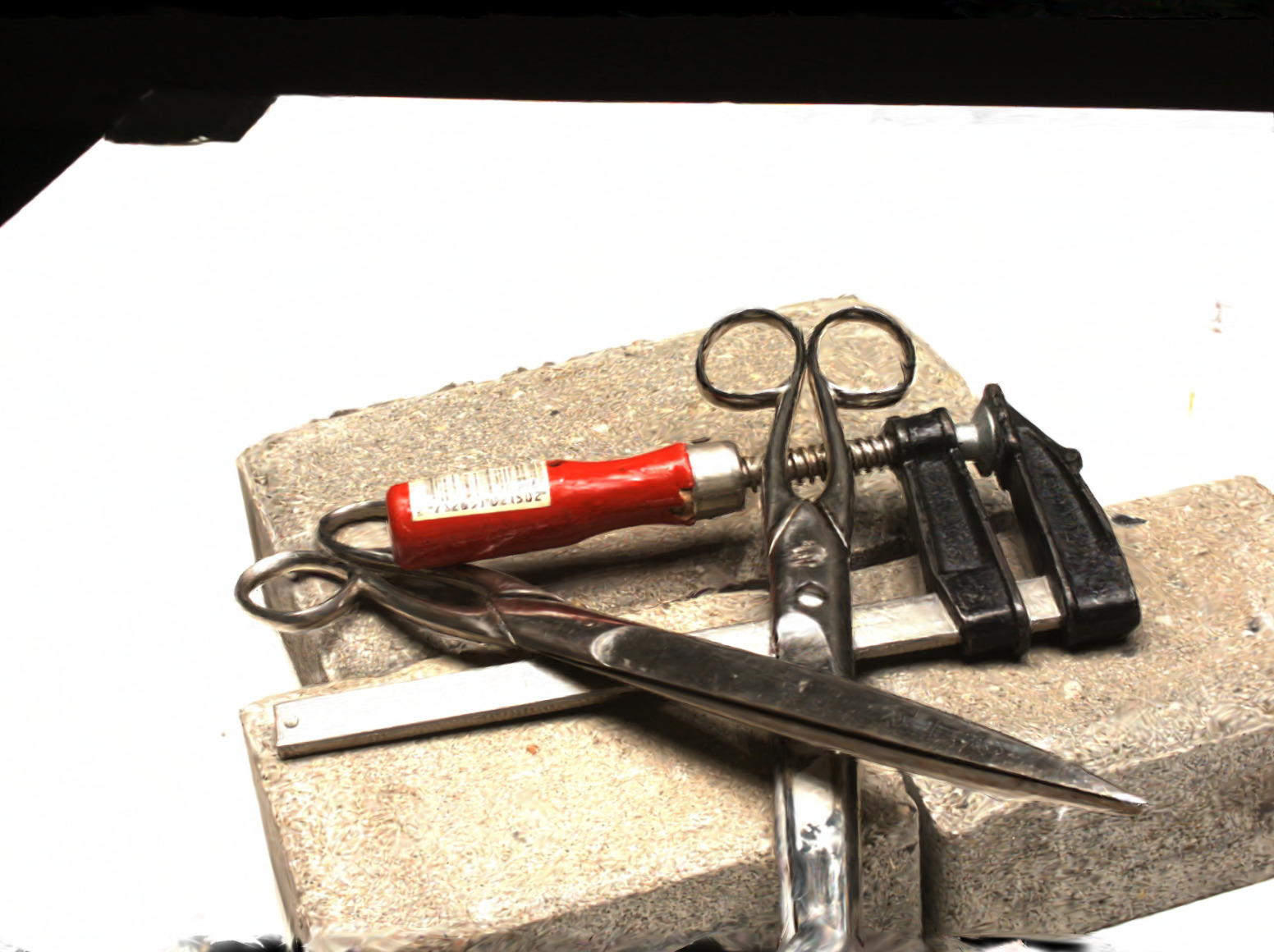}
        \includegraphics[width=0.98\linewidth]{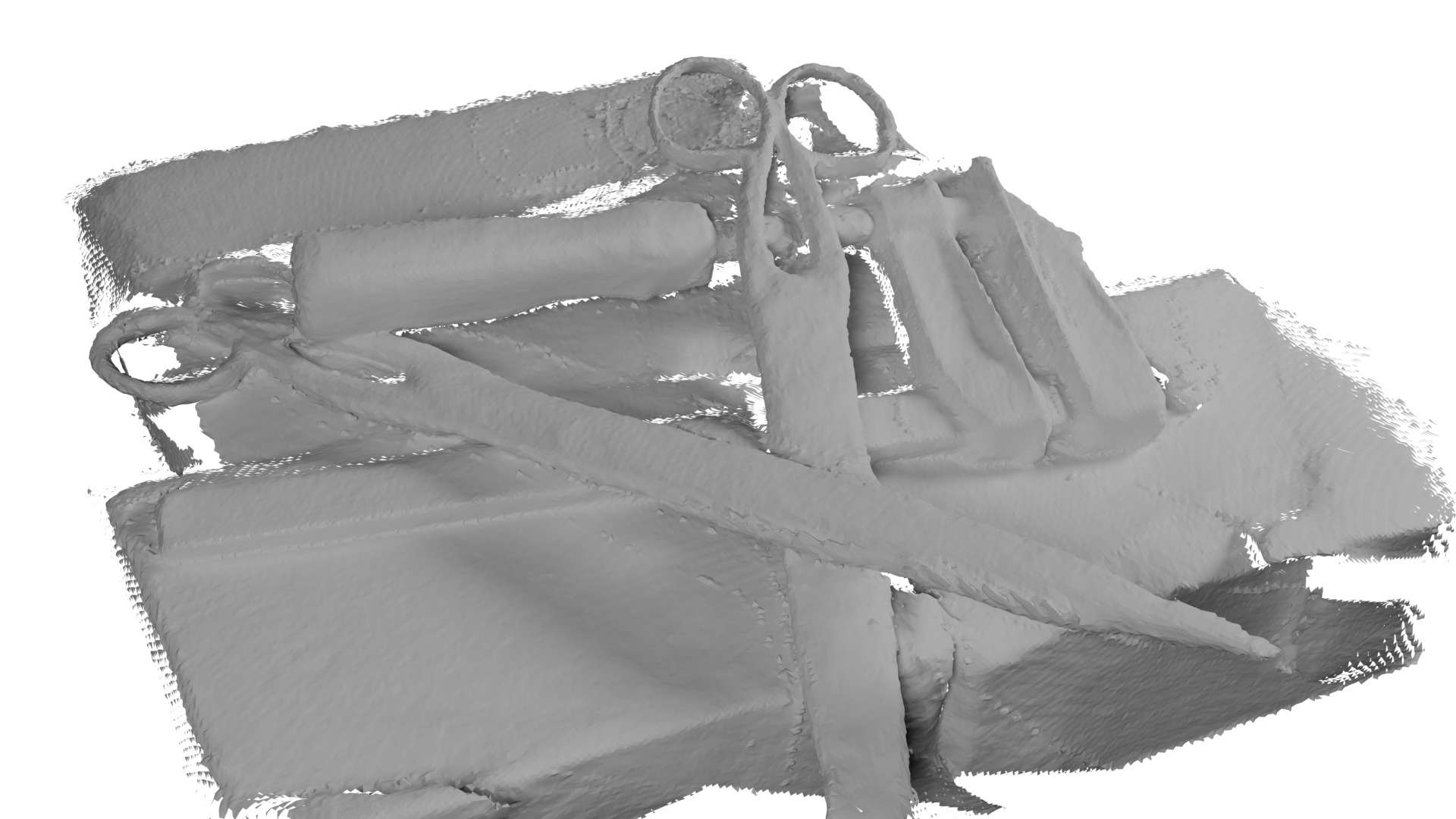}
        \vfill
        3 views
    \end{minipage}
    \begin{minipage}[b]{0.16\textwidth}
        \centering
        \includegraphics[trim={2.2cm 0 3cm 0},clip,width=0.98\linewidth]{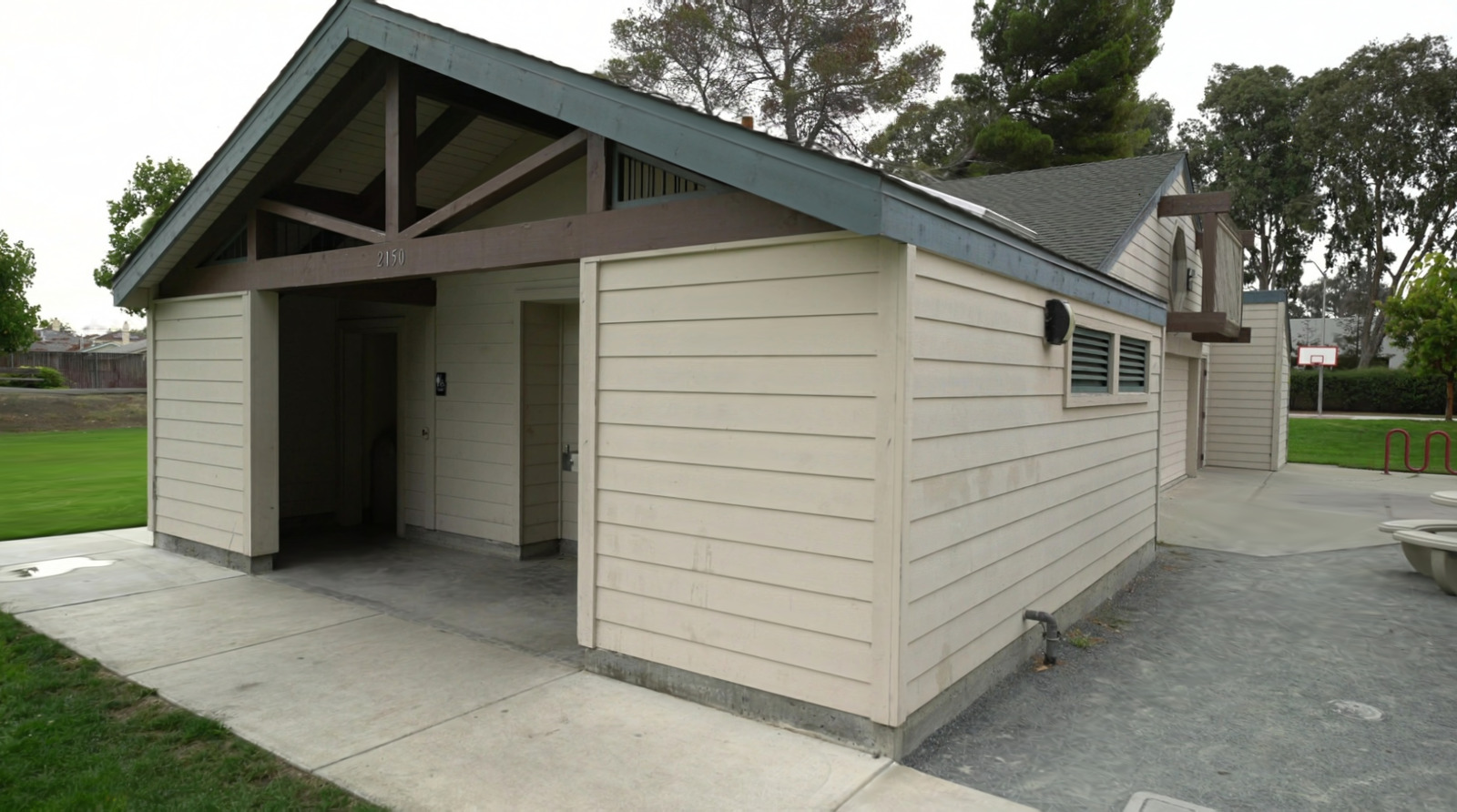}
        \includegraphics[width=0.98\linewidth]{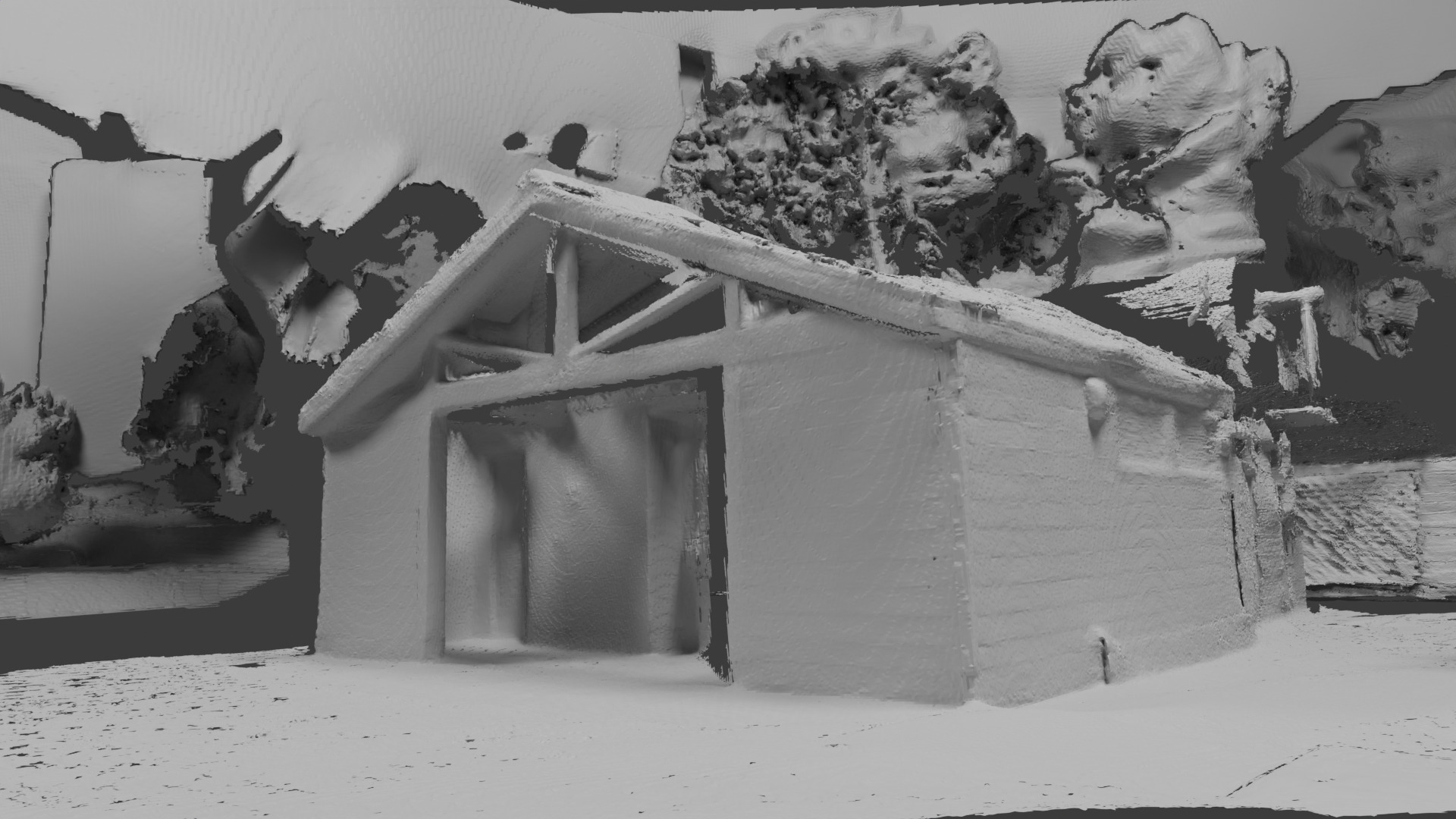}
        \vfill
        5 views
    \end{minipage}
    \begin{minipage}[b]{0.16\textwidth}
        \centering
        \includegraphics[trim={0 0.8cm 0 0},clip,width=0.98\linewidth]{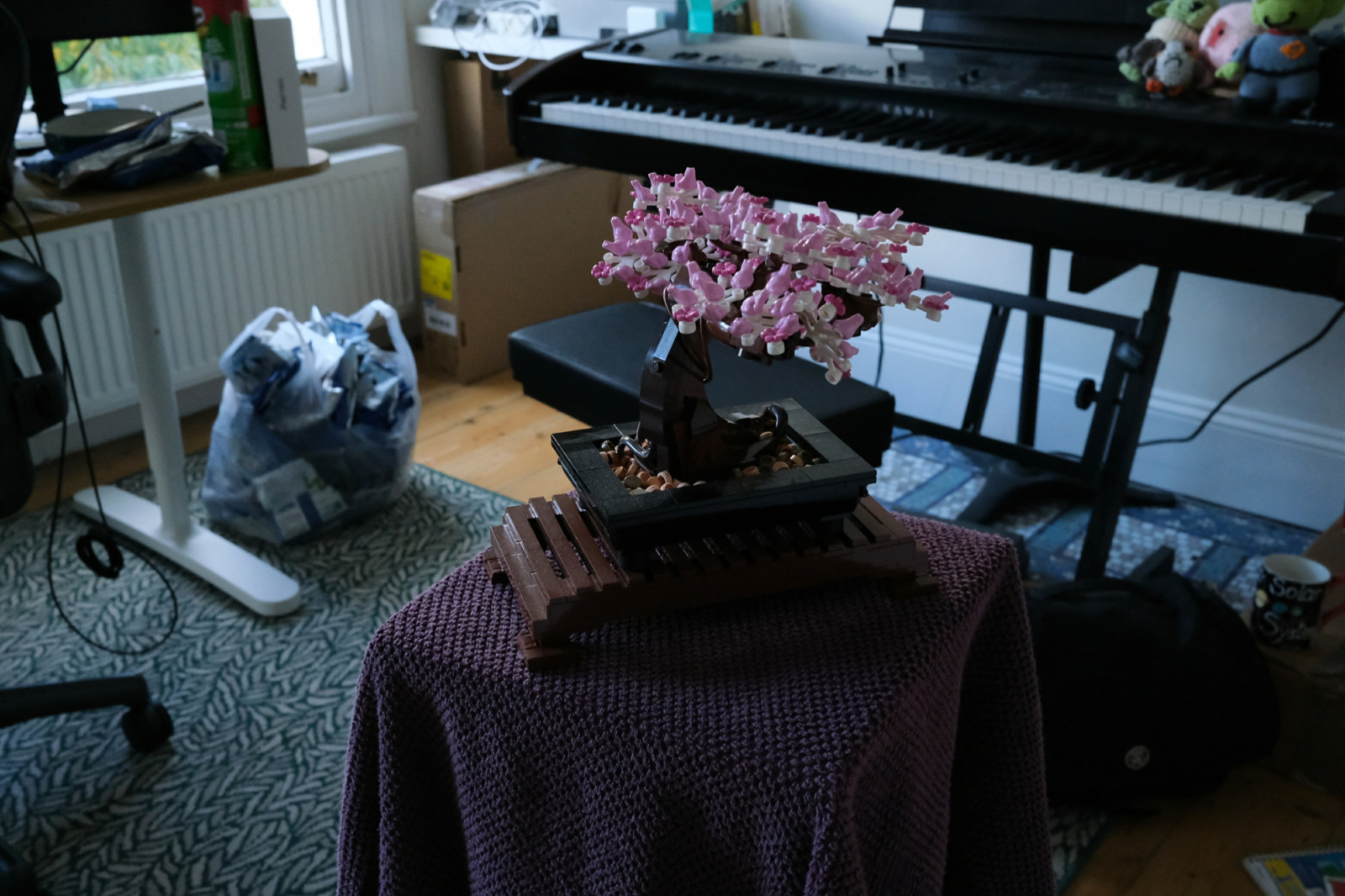}
        \includegraphics[width=0.98\linewidth]{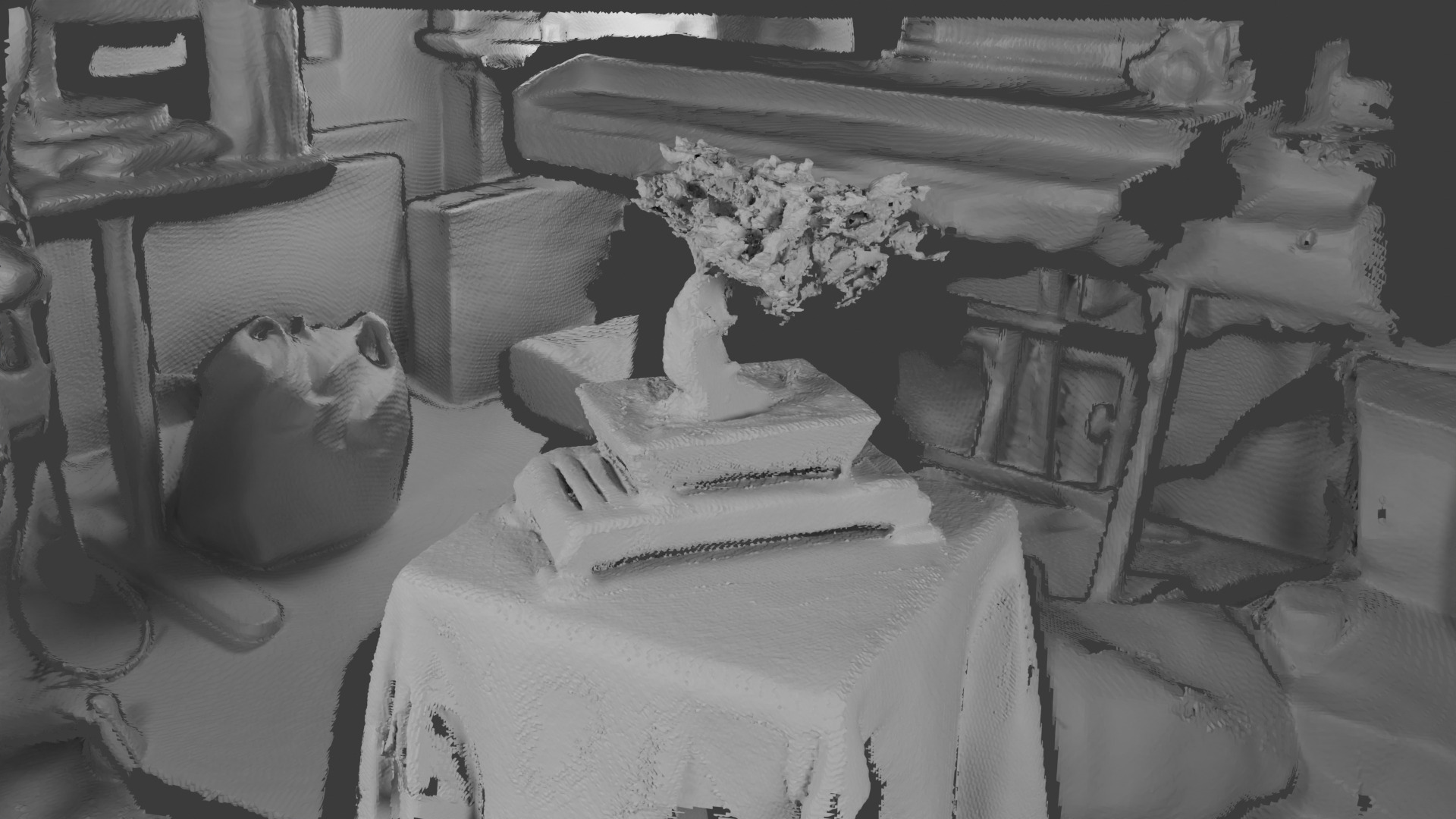}
        \vfill
        5 views
    \end{minipage}
    \begin{minipage}[b]{0.16\textwidth}
        \centering
        \includegraphics[trim={0 0.6cm 0 0},clip,width=0.98\linewidth]{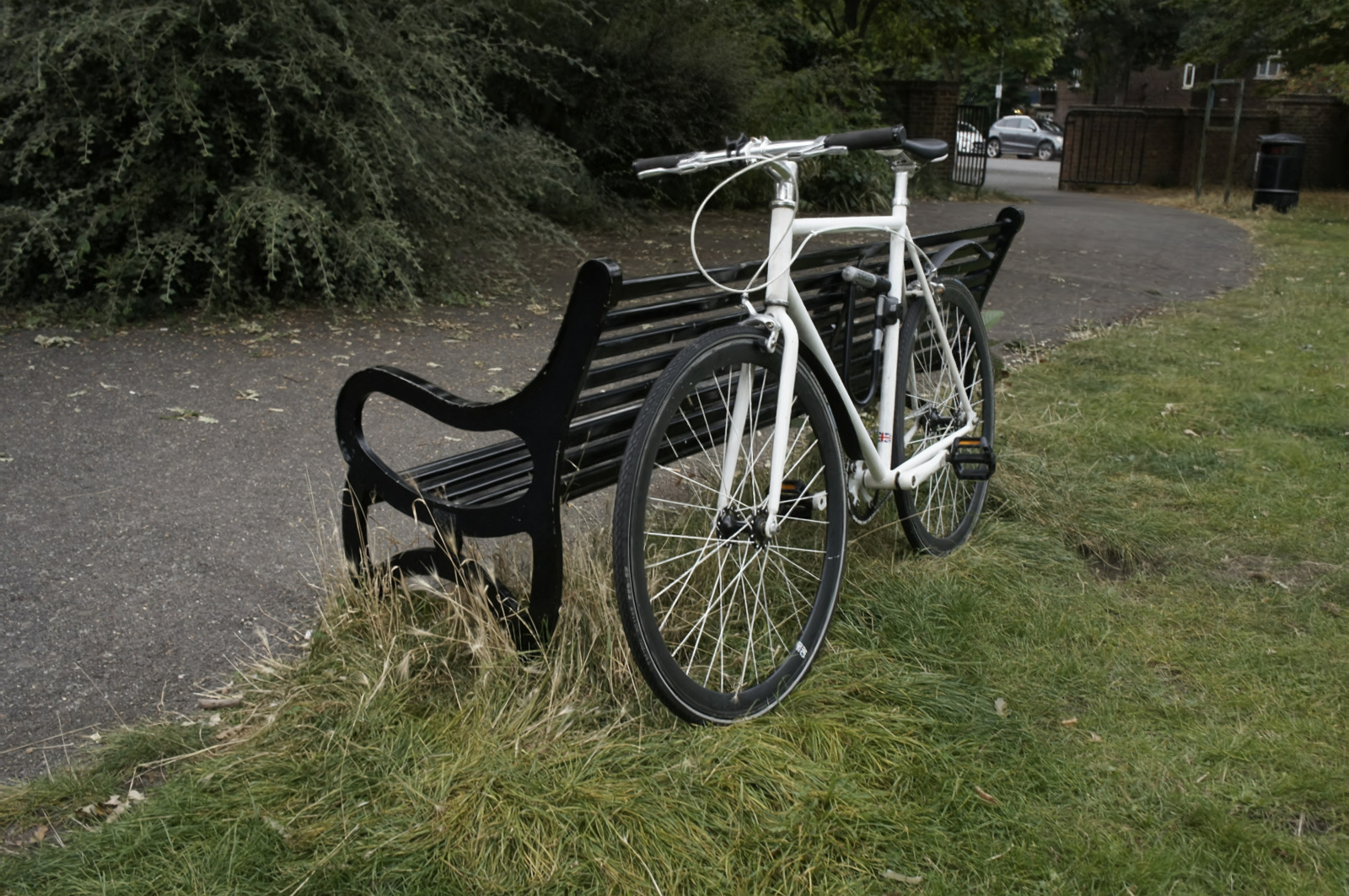}
        \includegraphics[width=0.98\linewidth]{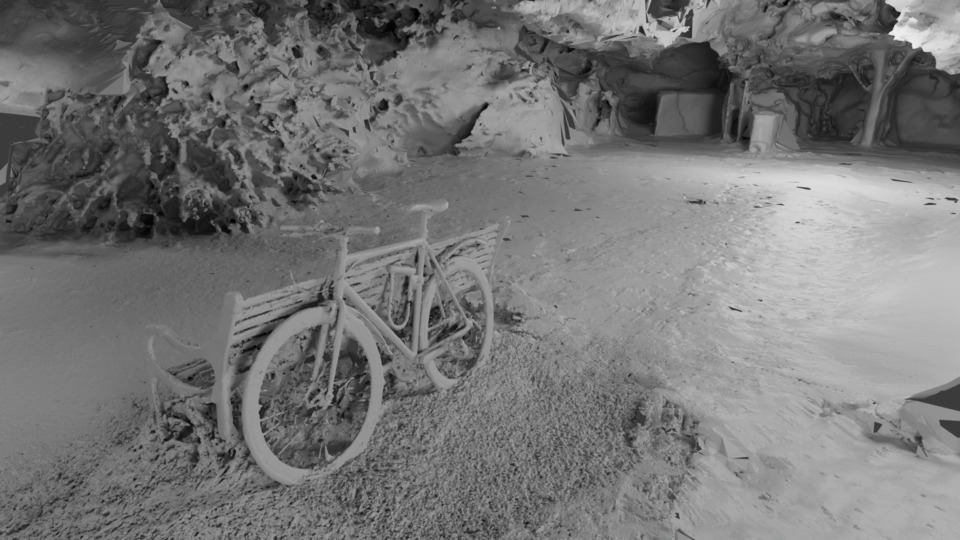}
        \vfill
        10 views
    \end{minipage}
    \begin{minipage}[b]{0.16\textwidth}
        \centering
        \includegraphics[trim={0 0.58cm 0 0},clip,width=0.98\linewidth]{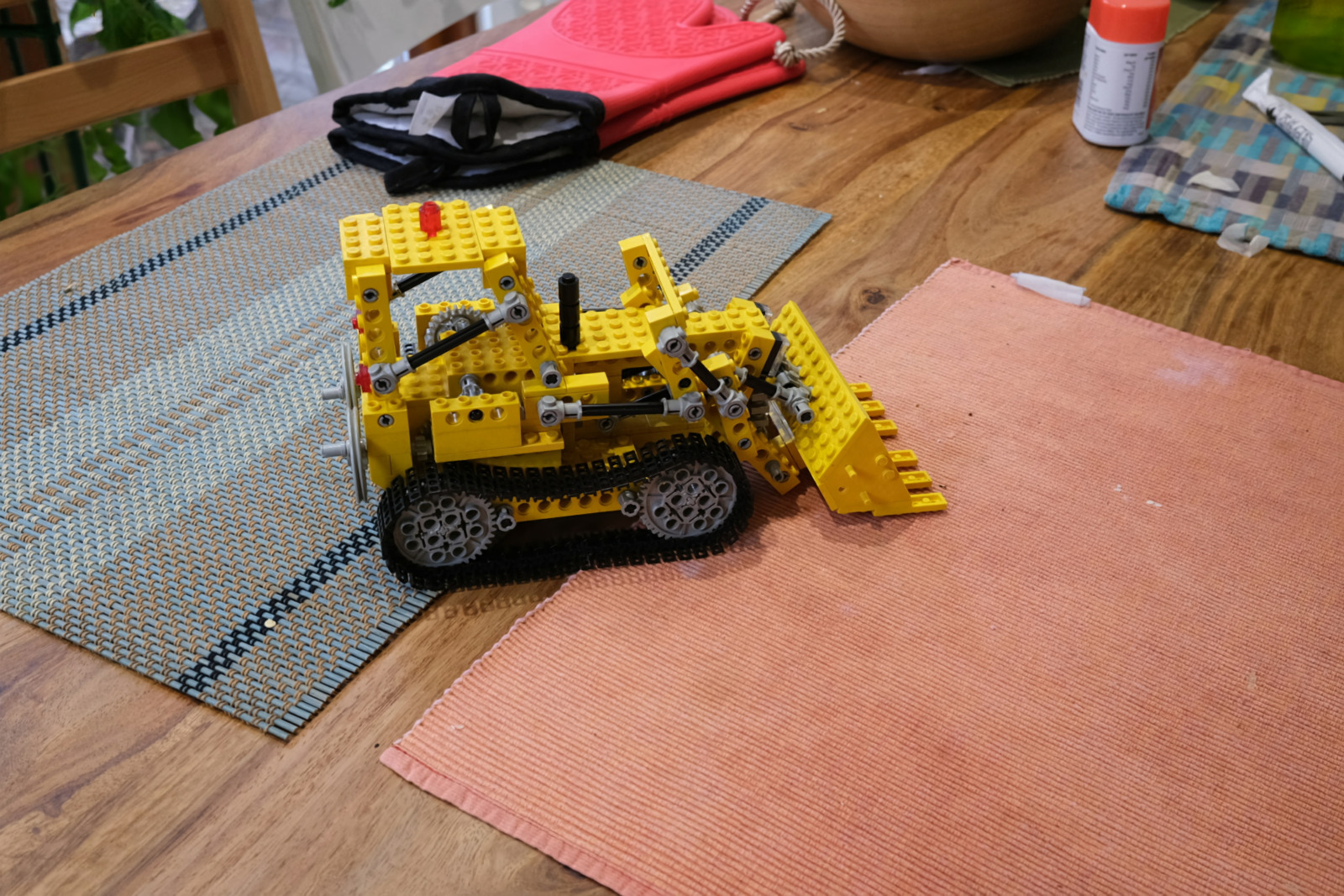}
        \includegraphics[width=0.98\linewidth]{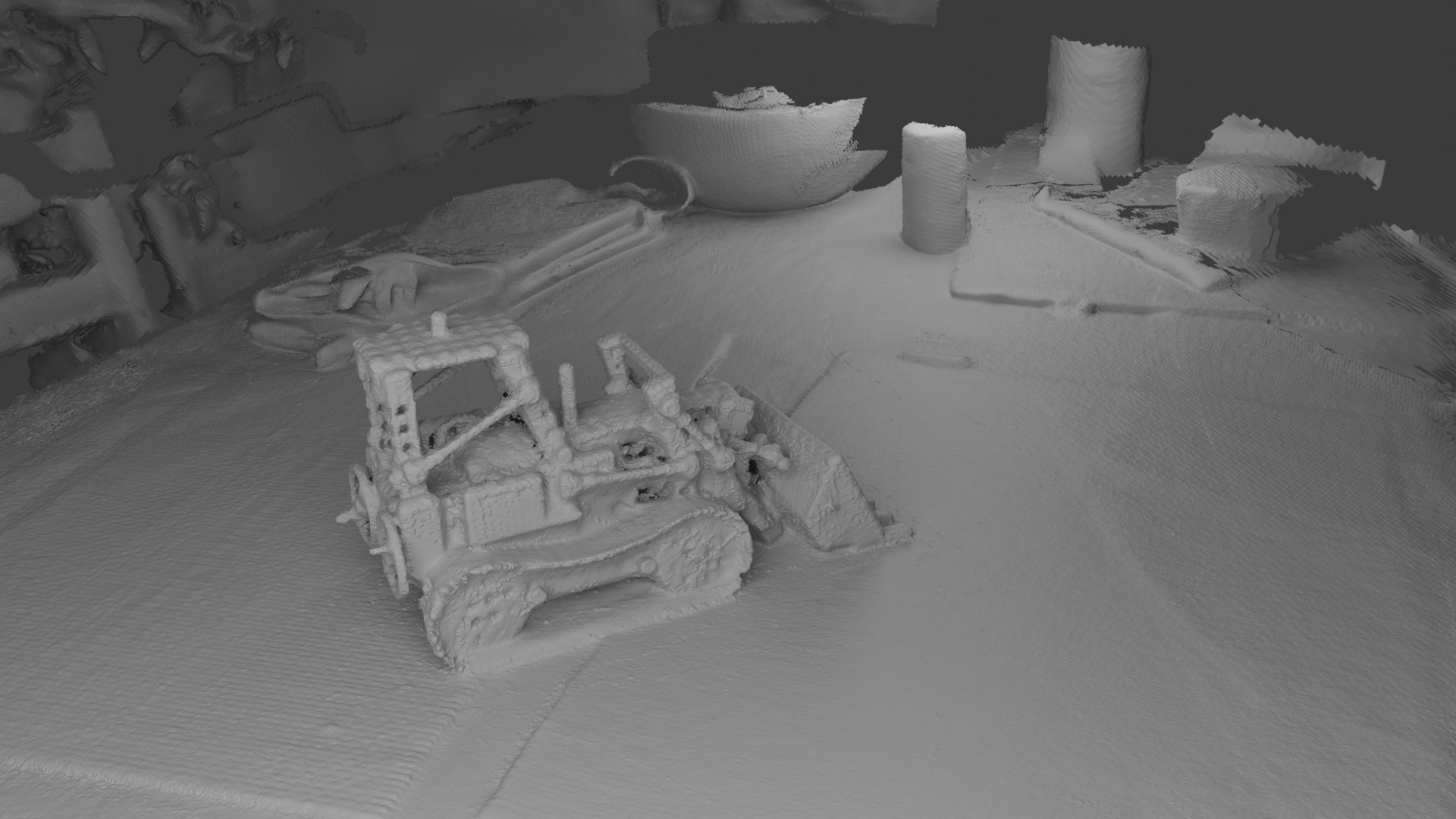}
        \vfill
        10 views
    \end{minipage}
    
    \caption{\textbf{Reconstruction with different numbers of input views.} Our method can produce high-quality renderings (top) and surfaces (bottom) even with very sparse input views (3-10 views). The quality of our meshes is visually pleasing even in extreme sparse scenarios.\tocheck{ADD ALSO NORMALS and maybe depth. Our normals are excellent, so we should showcase them.}}
    \label{fig:different_number_of_views}
\end{figure*}

\subsection{Refining the Manifold with Gaussian Surfels}

We refine our manifold representation with a photometric rendering loss. 
We instantiate 2D Gaussian surfels on the fly to texture our charts and render them with a Gaussian surfel rasterizer~\cite{huang20242d}.
Specifically, we learn color and opacity textures for each chart.
Since we know the UV coordinates of any Gaussian we instantiate on the charts, we can use our textures to compute color and opacity values for all Gaussians. All other parameters of the Gaussians, such as positions and covariances, are not learnable and computed on the fly depending on the position of the vertices.

We use Gaussian surfels, instead of triangle rasterization, because once splat in the screen space, the support of a rasterized Gaussian is larger than the visible ellipsoid and covers neighboring pixels. Rendering charts with Gaussians realizes better propagation of gradients across the different pixels, in contrast to triangle rasterization which would require blurring on the rendering to help propagate gradients. 
In this regard, Gaussian surfels could be considered as local kernels performing adaptive Gaussian blurring dependent on the size of the triangles.

We exploit a conventional photometric loss~\cite{kerbl3Dgaussians}, a regularization term from 2DGS~\cite{huang20242d}, and the structure loss for this refinement. We weight the structure term using our confidence maps $C_i$ for depth regularization robust to outliers. Please refer to the appendix for more details.
After refinement, we can extract a single-piece mesh from our manifold. 

\subsection{Extracting Meshes from Gaussian Surfels}

\begin{figure}[t]
    \centering
    \vspace{12pt}
    {\small
    \begin{minipage}[b]{0.23\textwidth}
        \centering
        \includegraphics[width=1.\linewidth]{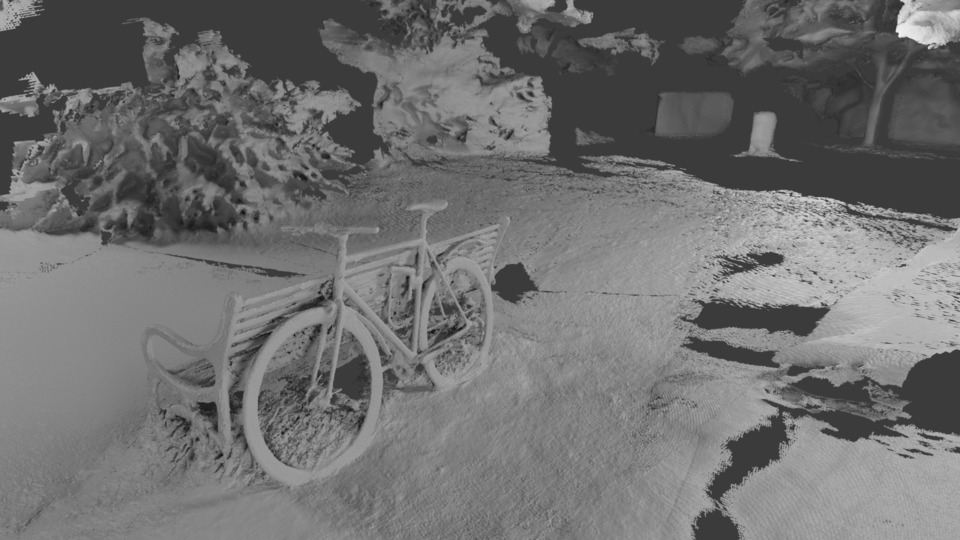}
        \vfill
        \includegraphics[width=1.\linewidth]{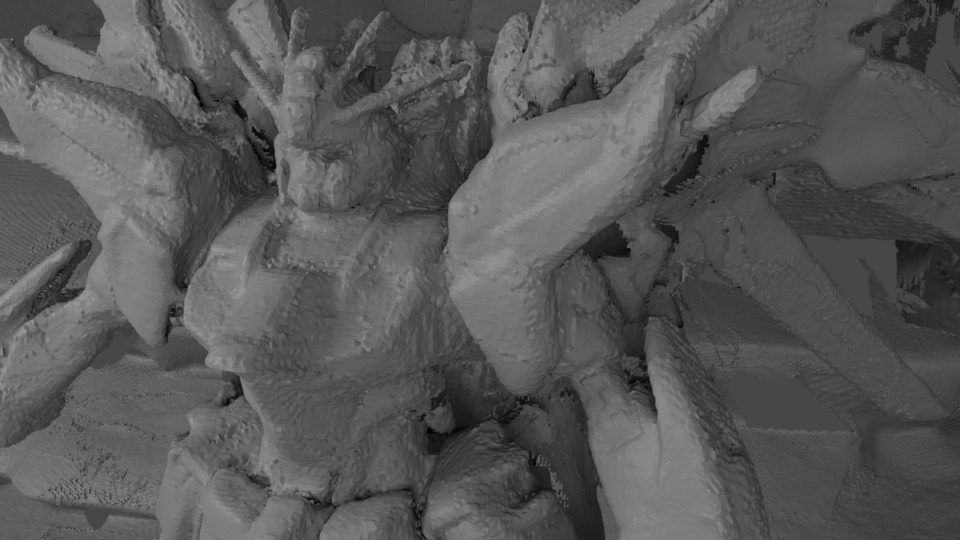}
        \vfill
        Multi-resolution TSDF
    \end{minipage}
    \begin{minipage}[b]{0.23\textwidth}
        \centering
        \includegraphics[width=1.\linewidth]{images/mesh_extraction/adaptive_tetra.jpg}
        \vfill
        \includegraphics[width=1.\linewidth]{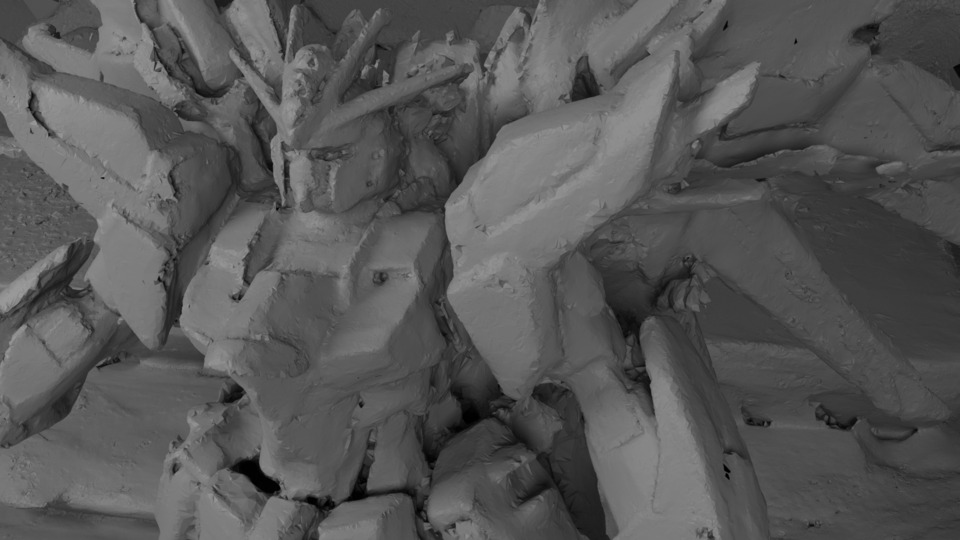}
        \vfill
        Adaptive tetrahedralization
    \end{minipage}
    }
    
    \caption{
    \textbf{Comparison between our two different mesh extraction methods: Multi-resolution TSDF fusion (left), and Adaptive tetrahedralization (right).} We optimized MAtCha~Gaussians representations with only 10 training images.
    Contrary to vanilla TSDF fusion, our multi-resolution TSDF can reconstruct both foreground and background objects with a decent number of vertices. However, similarly to vanilla TSDF fusion, it produces eroded meshes with holes in the surface, as well as ``\emph{disk-aliasing}'' artifacts. On the contrary, our adaptive tetrahedralization inspired by GOF~\cite{yu2024gaussian} is able to reconstruct accurate and complete surfaces meshes (see top right image), with sharp and fine details (see bottom right image).
    }
    \label{fig:mesh-extraction}
\end{figure}

Most existing surface reconstruction methods relying on 3D Gaussians or Gaussian Surfels~\cite{huang20242d,dai2024high} apply TSDF~fusion on rendered depth maps to extract a mesh from the volumetric representation.
However, as observed in~\cite{yu2024gaussian}, TSDF~fusion is limited to bounded scenes and does not allow for extracting high-quality meshes including both foreground and background objects of the scene. 
Moreover, applying TSDF~fusion on Gaussian Surfels can over-smooth the geometry, erode fine details, and produce artifacts, such as ``\emph{disk-aliasing}'' patterns on the surface.
To solve this issue and generate meshes with better quality, Gaussian~Opacity~Fields~\cite{yu2024gaussian} proposes to use an adaptive tetrahedralization to extract a surface mesh with optimal resolution, where the local density of the SDF grid depends on the positions and the size of the closest Gaussians. The SDF values used during the tetrahedralization are derived from an opacity field, which is defined using 3D Gaussians and GOF~\cite{yu2024gaussian} rasterization. 
Because the opacity field relies on GOF's specificities to be computed, the tetrahedralization cannot be directly applied on different Gaussian-based representations, such as Gaussian surfels.

In this regard, while we propose a custom multi-resolution TSDF~fusion including foreground and background objects in our implementation, we also propose to adapt the tetrahedralization from GOF~\cite{yu2024gaussian} to make it compatible with any Gaussian-based method capable of rendering perspective-accurate depth maps.

First, we propose to change the definition of the opacity field, using depth maps instead of 3D Gaussians as in GOF: For any set of input depth maps, we define a binary opacity field from the depth maps as well as an adaptive dilation operation to avoid eroding geometry during mesh extraction.
Second, because the tetrahedralization introduced in GOF generally produces very large meshes with more than 10M vertices, we propose a new sampling strategy to build the initial tetrahedron grid to easily adjust or lower the resolution of the output mesh.
Please refer to our implementation for more details. A qualitative comparison of mesh extraction methods is available in \cref{fig:mesh-extraction}.

We believe that our adaptation of GOF~tetrahedralization provides a high-quality alternative to TSDF~fusion and can generalize to most Gaussian-based surface reconstruction methods.
\section{Experiments}

We focus on two different tasks to thoroughly evaluate the effectiveness of \method: surface mesh reconstruction from sparse RGB images, and novel view synthesis from sparse RGB images. We evaluate our method in both bounded and unbounded environments.

\paragraph{Implementation Details}
We implement our method in PyTorch and conduct experiments on a single NVIDIA RTX A6000 GPU. For all experiments, we initialize our manifold charts using DepthAnythingV2~\cite{depth_anything_v2} as the monodepth model, and use MASt3R-SfM~\cite{duisterhof2024mast3rsfm} for camera pose estimation in sparse-view scenarios. 
For reconstructing high-quality surfaces from 3 to 10 input views, our pipeline takes less than 3 minutes for aligning the charts, and between 5 and 10 minutes for refinement.

\paragraph{Surface Reconstruction}
\label{subsec:exp-surface}

\begin{table*}
  \centering
  {
  \scriptsize
  \begin{tabular}{@{}lccccccccccccc@{}}
    \toprule
     \textbf{Scan ID} & \textbf{21} & \textbf{24} & \textbf{34} & \textbf{37} & \textbf{38} & \textbf{40} & \textbf{82} & \textbf{106} & \textbf{110} & \textbf{114} & \textbf{118} & \textbf{Mean CD} \\
    \midrule
    Points2Surf~\cite{erler20points2surf} & 3.73 & 2.85 & 2.55 & 5.13 & 3.85 & 2.41 & 2.30 & 3.95 & 3.33 & 2.37 & 2.84 & 3.21 \\
    CAP-UDF~\cite{zhou2024cap-pami} & 2.72 & 1.53 & 1.45 & 4.05 & 2.78 & 1.81 & 4.22 & 3.51 & 3.83 & 2.24 & 3.65 & 2.89 \\
    \midrule
    NeuS~\cite{wang2021neus} & 4.52 & 3.33 & 3.03 & 4.77 & 1.87 & 4.35 & \cellcolor{yellow!25}1.89 & 4.18 & 5.46 & 1.09 & 2.40 & 3.36 \\
    VolSDF~\cite{wu23svolsdf} & 4.54 & 2.61 & 1.51 & 4.05 & \cellcolor{yellow!25}1.27 & 3.58 & 3.48 & 2.62 & 2.79 & \cellcolor{red!25}0.52 & 1.10 & 2.56 \\
    \midrule
    SuGaR~\cite{guedon2024sugar} & 2.71 & 2.04 & 2.14 & 4.01 & 2.90 & 2.45 & 4.68 & 3.82 & 3.28 & 2.44 & 2.66 & 3.01 \\
    2D Gaussian Splatting~\cite{huang20242d} + MASt3R-SfM~\cite{duisterhof2024mast3rsfm} & \cellcolor{orange!25}1.43 & \cellcolor{yellow!25}1.29 & 2.02 & 2.79 & 2.05 & \cellcolor{yellow!25}1.71 & 2.24 & \cellcolor{orange!25}1.23 & 2.26 & 0.85 & 1.72 & 1.79 \\
    Gaussian Opacity Fields~\cite{yu2024gaussian} + MASt3R-SfM~\cite{duisterhof2024mast3rsfm} & \cellcolor{yellow!25}1.71 & 1.37 & 1.41 & \cellcolor{orange!25}2.38 & 1.59 & 2.05 & 2.20 & 1.62 & 1.99 & 1.21 & 1.81 & \cellcolor{yellow!25}1.76\\
    \midrule
    SparseNeus~\cite{long22sparseneus} & 3.73 & 4.48 & 3.28 & 5.21 & 3.29 & 4.21 & 3.30 & 2.73 & 3.39 & 1.40 & 2.46 & 3.41 \\
    VolRecon~\cite{ren23volrecon} & 3.05 & 3.30 & 2.27 & 4.36 & 2.51 & 3.24 & 3.30 & 3.10 & 3.58 & 1.86 & 3.68 & 3.11 \\
    S-VolSDF~\cite{wu23svolsdf} & 3.18 & 2.95 & 2.19 & 3.40 & 2.30 & 2.69 & 2.69 & 1.60 & \cellcolor{yellow!25}1.48 & 1.21 & 1.16 & 2.26 \\
    NeuSurf~\cite{huang24neusurf} & 3.22 & 2.42 & \cellcolor{yellow!25}1.38 & 2.61 & 1.72 & 3.46 & 2.68 & 1.44 & 2.42 & \cellcolor{yellow!25}0.61 & \cellcolor{red!25}0.87 & 2.08 \\
    Spurfies~\cite{raj2024spurfies} & 2.36 & \cellcolor{orange!25}1.12 & \cellcolor{red!25}0.83 & \cellcolor{yellow!25}2.39 & \cellcolor{orange!25}1.14 & \cellcolor{orange!25}1.55 & \cellcolor{orange!25}1.67 & \cellcolor{yellow!25}1.26 & \cellcolor{orange!25}1.14 & \cellcolor{yellow!25}0.61 & \cellcolor{yellow!25}0.94 & \cellcolor{orange!25}1.36 \\
    \midrule
    \textbf{\method (Ours)} 
    \textbf{\method (Ours)} 
    & \cellcolor{red!25}\textbf{1.27}  
    & \cellcolor{red!25}\textbf{0.88}   
    & \cellcolor{orange!25}\textbf{0.85}  
    & \cellcolor{red!25}\textbf{1.89}  
    & \cellcolor{red!25}\textbf{1.08}  
    & \cellcolor{red!25}\textbf{1.06}  
    & \cellcolor{red!25}\textbf{1.15}  
    & \cellcolor{red!25}\textbf{0.89}  
    & \cellcolor{red!25}\textbf{0.87}  
    & \cellcolor{orange!25}\textbf{0.58}  
    & \cellcolor{orange!25}\textbf{0.89}  
    & \cellcolor{red!25}\textbf{1.04} \\  
    \bottomrule
  \end{tabular}
  }
  \caption{\textbf{Quantitative evaluation of surface reconstruction in a sparse-view scenario (3 images only), based on Chamfer~Distance~(mm)~($\downarrow$) on the DTU~dataset~\cite{aanaes2016large}.} We evaluate the quality of meshes reconstructed with various methods, using only 3 input RGB images. We outperform the previous best method Spurfies~\cite{raj2024spurfies} by~\textbf{24\%} on average~(\textbf{1.04} vs 1.36 CD). 
  {Moreover, our method partly relies on 2D Gaussian rasterization~\cite{huang20242d}, which requires cameras to have a centered principal point. As a consequence, we had to crop input images for extracting our meshes, resulting in incomplete reconstruction for some scenes such as scans 34 and 38, for instance. In this regard, even though we outperform previous works, the performance of our method is underestimated in this experimental setup.}
  }
  \label{tab:dtu_geometry}
\end{table*}
\begin{table}
  \centering
  {\scriptsize
  \begin{tabular}{@{}lccc@{}}
    \toprule
     & \textbf{5 training images} & \textbf{10 training images}\\
    \midrule
    2DGS~\cite{huang20242d}+MASt3R-SfM~\cite{duisterhof2024mast3rsfm} & \cellcolor{yellow!25}0.052 & \cellcolor{yellow!25}0.121 \\
    GOF~\cite{yu2024gaussian}+MASt3R-SfM~\cite{duisterhof2024mast3rsfm} & \cellcolor{orange!25}0.054 & \cellcolor{orange!25}0.144 \\
    \method~(Ours) & \cellcolor{red!25}\textbf{0.072} & \cellcolor{red!25}\textbf{0.156}\\
    \bottomrule
  \end{tabular}
  }
  \caption{\textbf{Quantitative evaluation for surface reconstruction in a sparse-view scenario for unbounded scenes of the Tanks\&Temples~dataset~\cite{Knapitsch2017}.}
  We evaluate our approach and two baselines 
  based on F-Score~($\uparrow$).
  The baselines combine recent surface-reconstruction methods~\cite{huang20242d, yu2024gaussian} augmented with MASt3R-SfM~\cite{duisterhof2024mast3rsfm} for greater robustness to sparse-view inputs. }
  \label{tab:tandt_metrics}
\end{table}
\begin{figure}[t]
    \centering
    \vspace{12pt}
    {\small
    \begin{minipage}[b]{0.23\textwidth}
        \centering
        \includegraphics[width=1.\linewidth, height=0.55\linewidth]{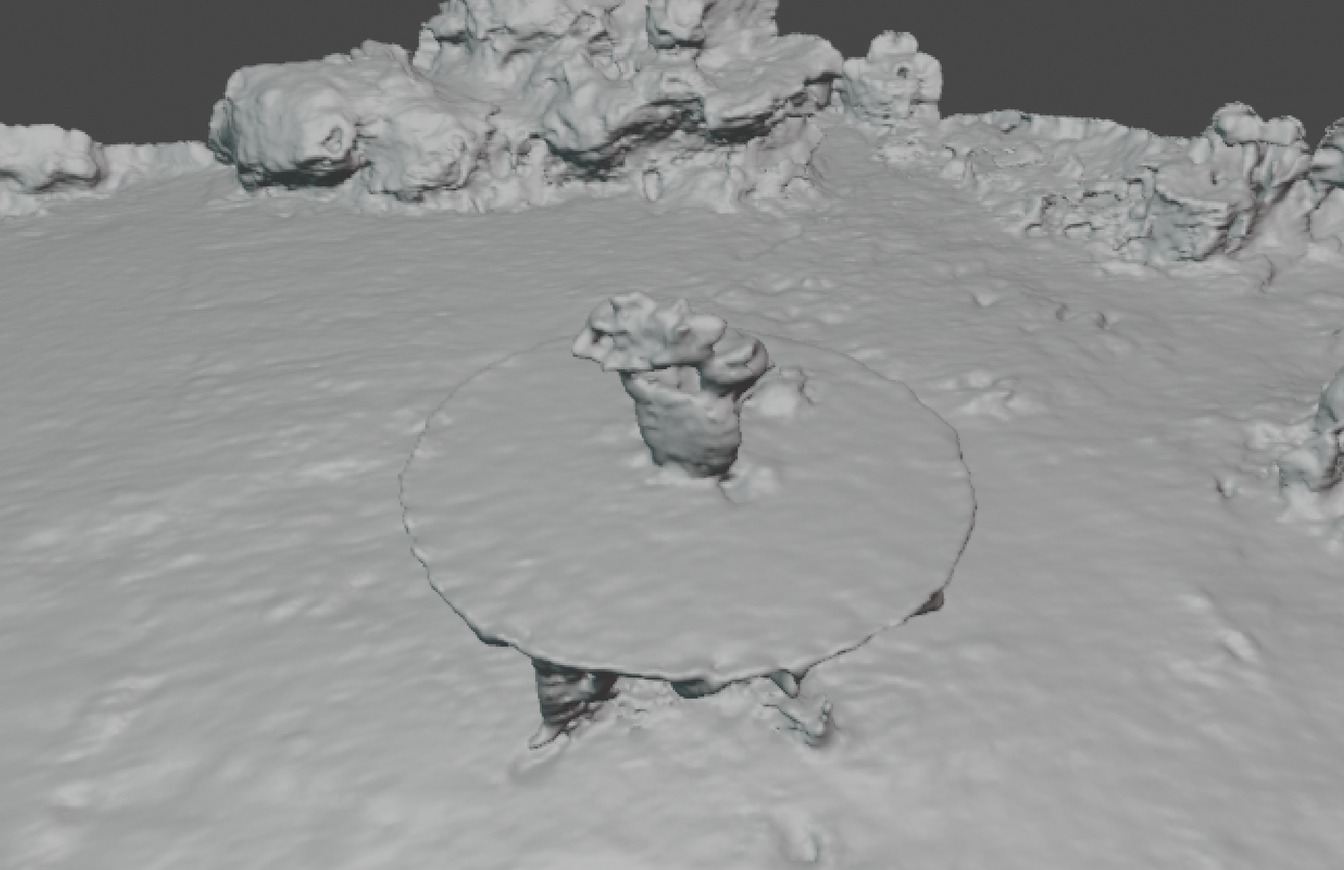}
        \vfill
        Spurfies~\cite{raj2024spurfies} \\
        \vfill
        \ \\
        \vfill
        \includegraphics[width=1.\linewidth, height=0.55\linewidth]{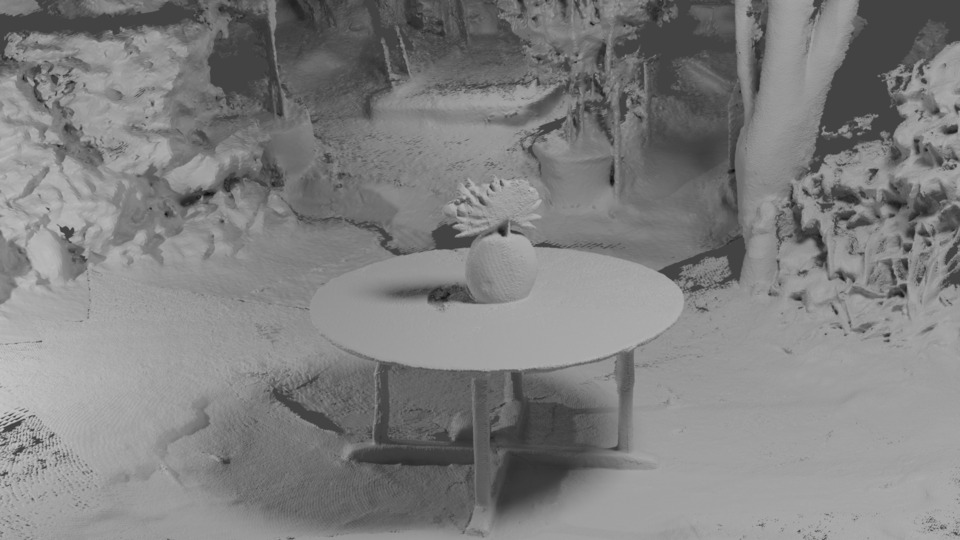}
        \vfill
        Ours
    \end{minipage}
    \begin{minipage}[b]{0.23\textwidth}
        \centering
        \includegraphics[width=1.\linewidth, height=0.55\linewidth]{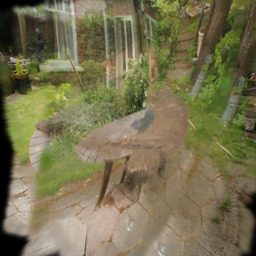}
        \vfill
        MVSplat~\cite{chen2024mvsplat} \\
        \vfill
        \ \\
        \vfill
        \includegraphics[trim={0 1.4cm 0 2.5cm},clip,width=1.\linewidth]{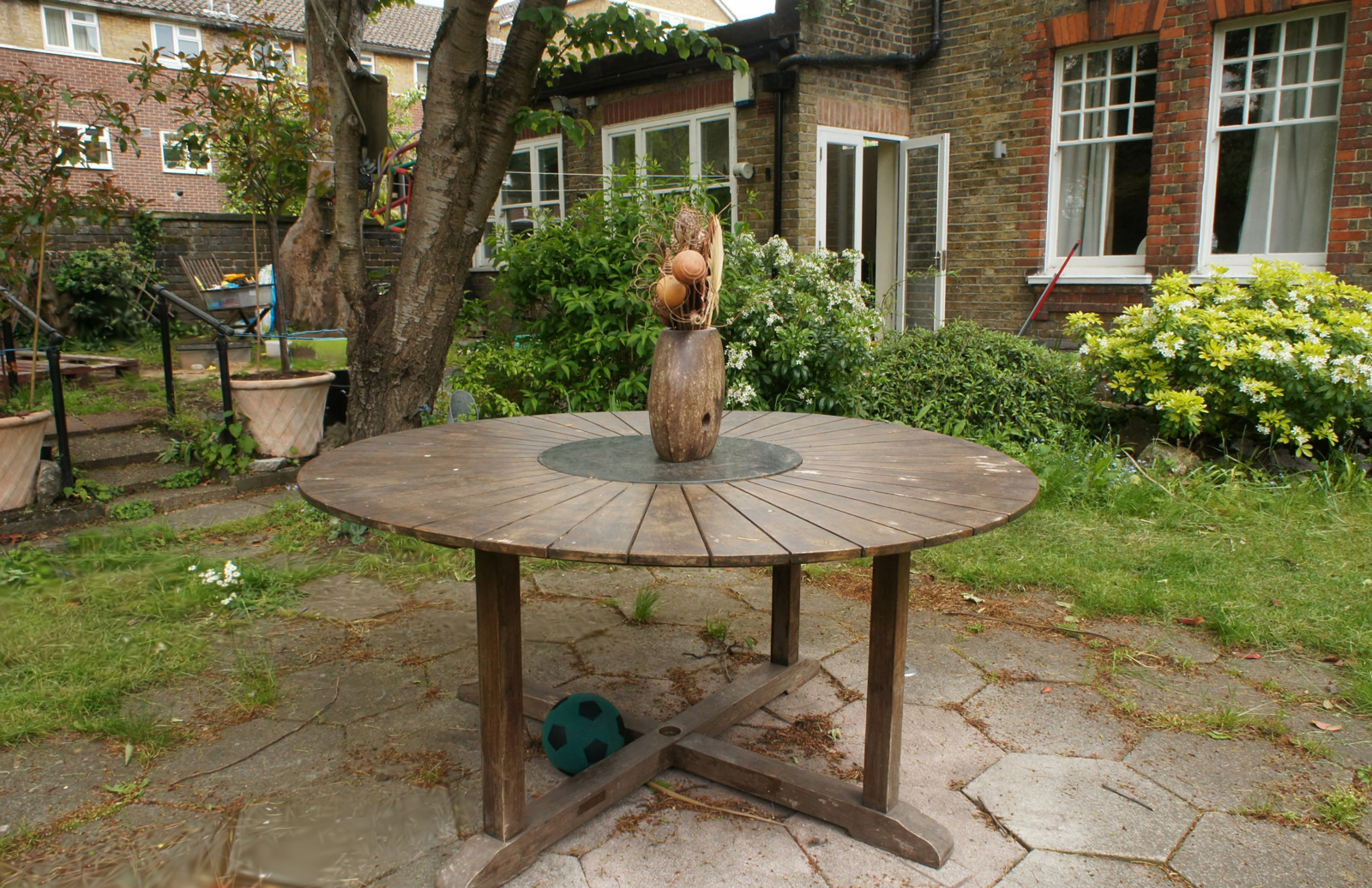}
        \vfill
        Ours
    \end{minipage}
    }
    
    \caption{\textbf{Comparisons with Spurfies~\cite{raj2024spurfies} 
    and MVSplat~\cite{chen2024mvsplat} 
    on an unbounded scene.} 
    Our method outperforms state-of-the-art approaches for surface reconstruction and feed-forward Gaussian splatting regression in sparse view scenarios.}
    \label{fig:comparison_with_SOTA}
\end{figure}

We evaluate the accuracy of our reconstructed meshes on two standard benchmarks: the DTU dataset~\cite{aanaes2016large} and Tanks~\&~Temples (T\&T)~dataset~\cite{Knapitsch2017}. For a fair comparison with prior work, in the experiments on the DTU dataset, we follow the evaluation protocol from previous sparse-view reconstruction methods Spurfies~\cite{raj2024spurfies} and S-VolSDF~\cite{wu23svolsdf}: We take three input views 22, 25, and 28 from each scan, use calibrated camera parameters for the global alignment of MASt3R-SfM~\cite{duisterhof2024mast3rsfm},
and filter the recovered meshes with masks before comparison.
For T\&T, we optimize the models using 5 and 10 images sparsely sampled in the scenes.
Since the implementation of Spurfies~\cite{raj2024spurfies} is not publicly available yet and none of other previous works evaluates sparse-view reconstruction in large or unbounded scenes, we propose two strong baselines for fair comparison on T\&T: We use MASt3R-SfM to initialize both 2DGS~\cite{huang20242d} and GOF~\cite{yu2024gaussian} and optimize the representations with depth-normal regularization. Indeed, augmenting recent state-of-the-art methods with MASt3R-SfM~\cite{duisterhof2024mast3rsfm} provides much better robustness in sparse-view scenarios.

\cref{tab:dtu_geometry,tab:tandt_metrics} show quantitative comparisons using Chamfer Distance (CD) and F-Score. 
Our method achieves state-of-the-art performance across both datasets, outperforming both traditional surface reconstruction approaches (Points2Surf~\cite{erler20points2surf}, 
CAP-UDF~\cite{zhou2024cap-pami}
) and recent neural methods (NeuS~\cite{wang2021neus}, VolSDF~\cite{wu23svolsdf}, Spurfies~\cite{raj2024spurfies}). Notably, we achieve these results with significantly faster reconstruction times---minutes versus hours for most baselines.

\cref{fig:different_number_of_views} shows qualitative results on different datasets~\cite{aanaes2016large,Knapitsch2017,barron22mipnerf360} with different numbers of views. The results show the effectiveness and robustness of our method in extreme sparse scenarios.

\vspace{-8pt}
\paragraph{Novel View Synthesis}

\begin{table*}
  \centering
  {\small
  \begin{tabular}{@{}lccccccc@{}}
    \toprule
     \multicolumn{1}{c}{} & \multicolumn{2}{c}{Mip-NeRF~360~\cite{barron22mipnerf360}} & \multicolumn{2}{c}{Tanks\&Temples~\cite{Knapitsch2017}} & \multicolumn{2}{c}{DeepBlending~\cite{hedman18deepblending}} \\
     \cmidrule(r){2-3} \cmidrule(r){4-5} \cmidrule(r){6-7}
      & 10\%Q PSNR $\uparrow$ & Avg PSNR $\uparrow$ & 10\%Q PSNR $\uparrow$ & Avg PSNR $\uparrow$ & 10\%Q PSNR $\uparrow$ & Avg PSNR $\uparrow$ \\
    \midrule
    \multicolumn{7}{l}{\textbf{5 training views}} \\
    \midrule
    2DGS~\cite{huang20242d}+MASt3R-SfM~\cite{duisterhof2024mast3rsfm} & \cellcolor{yellow!25}15.37 & \cellcolor{yellow!25}20.84 & \cellcolor{orange!25}14.23 & \cellcolor{yellow!25}16.42 & \cellcolor{orange!25}15.84 & \cellcolor{yellow!25}19.86  \\
    GOF~\cite{yu2024gaussian}+MASt3R-SfM~\cite{duisterhof2024mast3rsfm} & \cellcolor{orange!25}15.78 & \cellcolor{orange!25}21.24 & \cellcolor{yellow!25}13.69 & \cellcolor{orange!25}16.50 & \cellcolor{yellow!25}15.58 & \cellcolor{orange!25}19.87 \\
    \textbf{\method~(Ours)} & \cellcolor{red!25}\textbf{18.18} & \cellcolor{red!25}\textbf{21.90} & \cellcolor{red!25}\textbf{15.33} & \cellcolor{red!25}\textbf{17.30} & \cellcolor{red!25}\textbf{17.22} & \cellcolor{red!25}\textbf{20.60} \\
    \midrule
    \multicolumn{7}{l}{\textbf{10 training views}} \\
    \midrule
    2DGS~\cite{huang20242d}+MASt3R-SfM~\cite{duisterhof2024mast3rsfm} & \cellcolor{yellow!25}19.94 & \cellcolor{yellow!25}24.31 & \cellcolor{yellow!25}16.63 & \cellcolor{orange!25}19.59 & \cellcolor{orange!25}14.06 & \cellcolor{orange!25}21.14 \\
    GOF~\cite{yu2024gaussian}+MASt3R-SfM~\cite{duisterhof2024mast3rsfm} & \cellcolor{orange!25}20.99 & \cellcolor{orange!25}24.50 & \cellcolor{orange!25}16.81 & \cellcolor{orange!25}19.59 & \cellcolor{yellow!25}12.61 & \cellcolor{yellow!25}21.12 \\
    \textbf{\method~(Ours)} & \cellcolor{red!25}\textbf{21.55} & \cellcolor{red!25}\textbf{25.10} & \cellcolor{red!25}\textbf{17.96} & \cellcolor{red!25}\textbf{20.38} & \cellcolor{red!25}\textbf{17.41} & \cellcolor{red!25}\textbf{22.98} \\
    \bottomrule
  \end{tabular}
  }
  \caption{\textbf{Quantitative evaluation of Novel View Synthesis in sparse-view scenarios across multiple real-world datasets.} We evaluate our method against baselines on three challenging datasets: Mip-NeRF~360~\cite{barron22mipnerf360}, Tanks\&Temples~\cite{Knapitsch2017}, and DeepBlending~\cite{hedman18deepblending}. 
  Baselines consist of recent state-of-the-art approaches augmented with MASt3R-SfM~\cite{duisterhof2024mast3rsfm} for more robustness to sparse-view scenarios. 
  For each dataset and method, we report both the average PSNR and the 10\% quantile PSNR (10\%Q PSNR) which better reflects performance on challenging views and better capture the ability of a method to generalize to novel viewpoints. 
  Results are shown for both 5-view and 10-view scenarios, demonstrating our method's superior performance across different sparsity levels.}
  \label{tab:mipzipdb_nvs}
\end{table*}
\begin{figure}[t]
    \centering
    {\small
    \begin{minipage}[b]{0.15\textwidth}
        \centering
        \includegraphics[width=1.\linewidth]{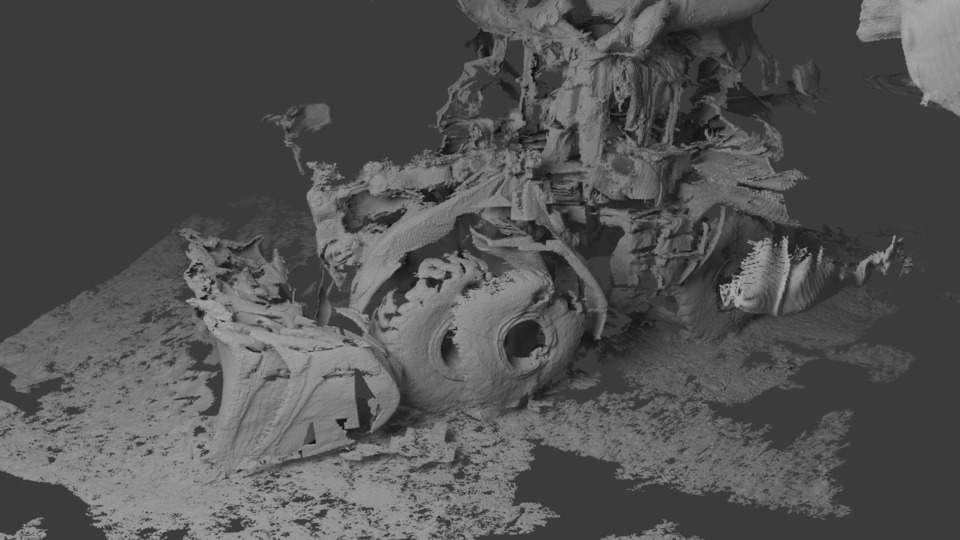}
        \vfill
        \includegraphics[width=1.\linewidth]{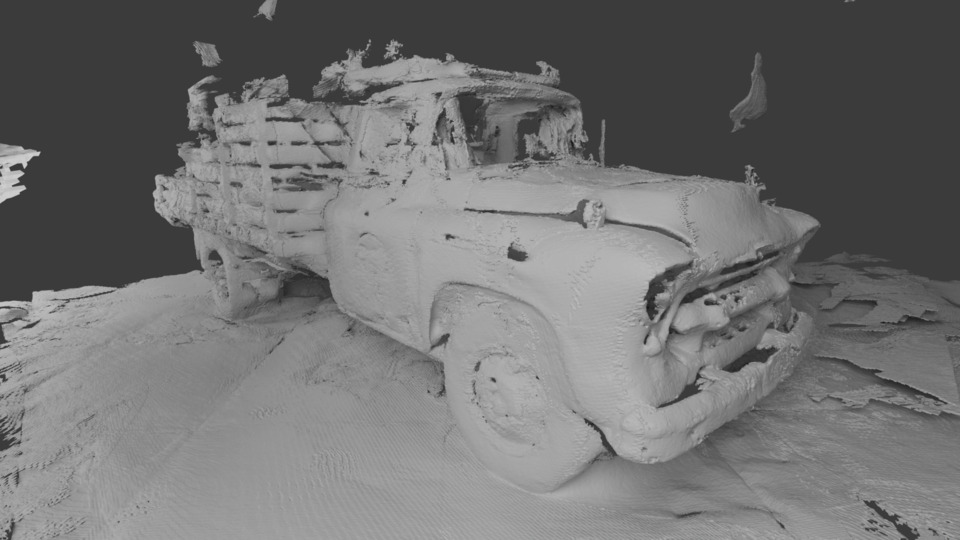}
        \vfill
        2DGS~\cite{huang20242d}\\+MASt3R-SfM~\cite{duisterhof2024mast3rsfm}
    \end{minipage}
    \begin{minipage}[b]{0.15\textwidth}
        \centering
        \includegraphics[width=1.\linewidth]{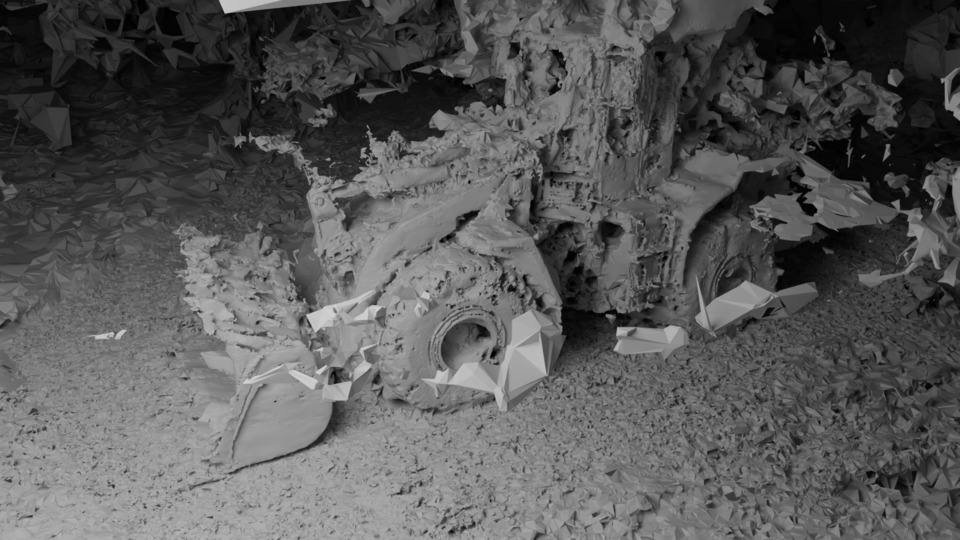}
        \vfill
        \includegraphics[width=1.\linewidth]{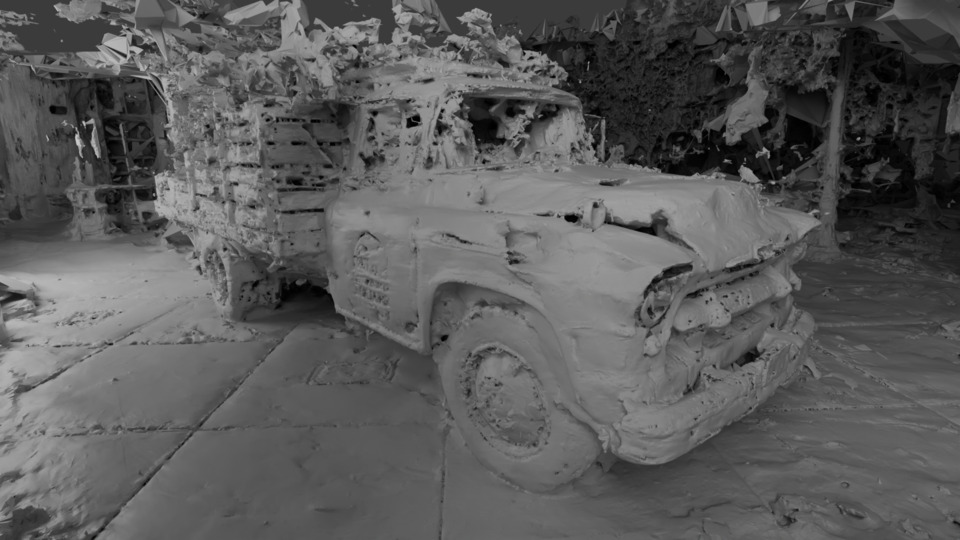}
        \vfill
        GOF~\cite{yu2024gaussian}\\+MASt3R-SfM~\cite{duisterhof2024mast3rsfm}
    \end{minipage}
    \begin{minipage}[b]{0.15\textwidth}
        \centering
        \includegraphics[width=1.\linewidth]{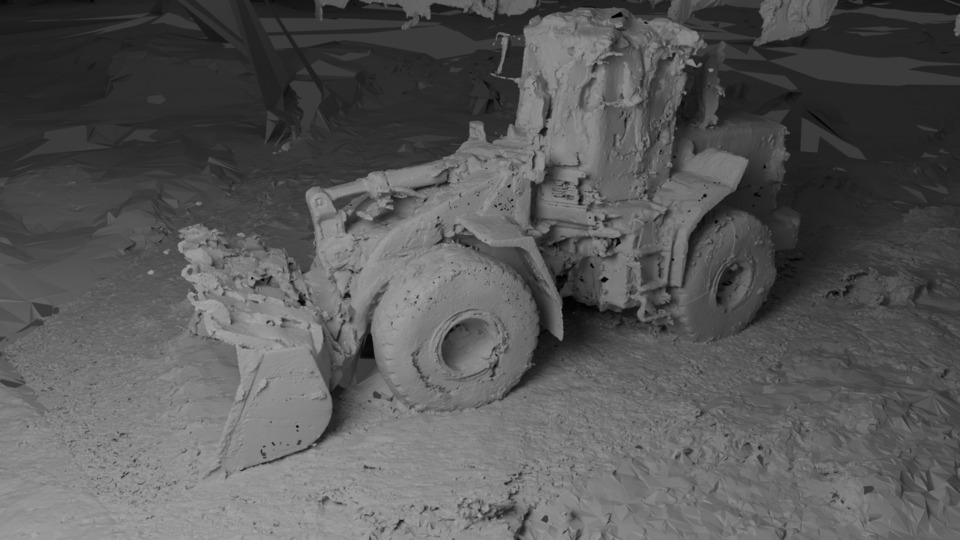}
        \vfill
        \includegraphics[width=1.\linewidth]{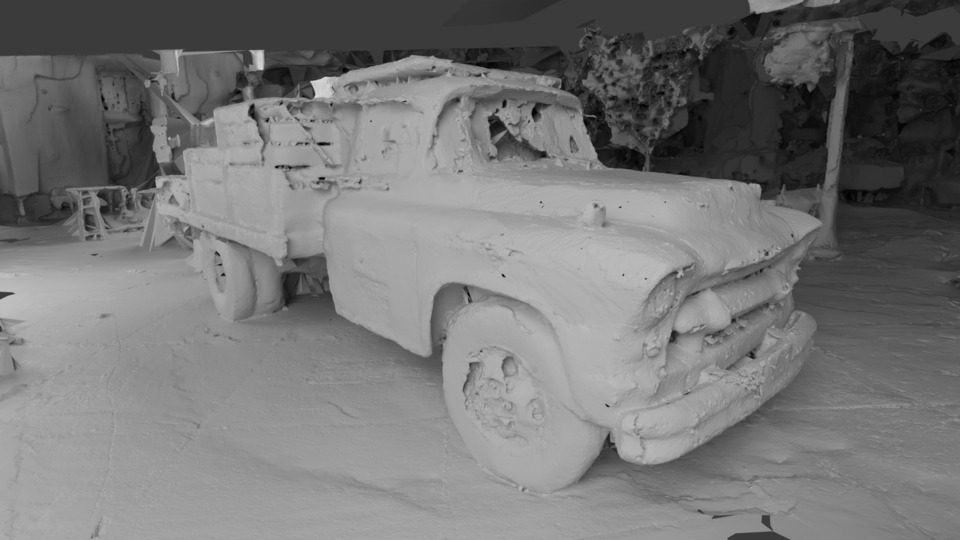}
        \vfill
        \method\\(Ours)
    \end{minipage}
    }
    
    \caption{
    \textbf{Qualitative evaluation for surface reconstruction in a sparse-view scenario for unbounded scenes from Tanks\&Temples~\cite{Knapitsch2017}, with 5 training images (top row) and 10 training images (bottom row).} 
    Contrary to the baselines, our approach is able to reconstruct accurate and complete surfaces meshes: It not only includes both foreground and background objects, but also recover sharper and finer details.
    \vspace{-12pt}
    }
    \label{fig:mesh-comparison}
\end{figure}

We provide results of novel view synthesis in sparse-view settings across three challenging real-world datasets of unbounded scenes: Mip-NeRF 360~\cite{barron22mipnerf360}, Tanks\&Temples~\cite{Knapitsch2017}, and DeepBlending~\cite{hedman18deepblending}.
Similarly to the previous section, we trained our models with either 5 or 10 input images, and evaluate them with 10 test images.
More details about the experimental setting are available in the appendix.
We report in \cref{tab:mipzipdb_nvs} both average PSNR and 10\% quantile PSNR (10\%Q PSNR) metrics. 
\kohei{The 10\%Q PSNR is the PSNR value below which 10\% of the test views fall. Average PSNR provides an overall measure of reconstruction quality, whereas the 10\%Q PSNR specifically captures accuracy on the most challenging views as well as the ability of a method to generalize to novel viewpoints.}
This metric is particularly relevant to sparse-view settings where some novel viewpoints may have very limited overlap with input views.

As shown in \cref{tab:mipzipdb_nvs} and \cref{fig:mesh-comparison}, our method consistently outperforms the baselines across all datasets and metrics even though its main focus is high-quality surface reconstruction, not novel view synthesis. 
In the 5-view scenario, we achieve significant improvements over the baselines. The performance gap remains substantial even when increasing to 10 input views, where our method maintains superior reconstruction quality across datasets. This consistent performance advantage demonstrates the effectiveness of our chart-based representation and refinement approach in handling sparse-view scenarios.
Note that test images include images with very little overlap with training images, which should explain the low values as well as the low variance in the results between the methods.

\begin{table}
  \centering
  {
  \small
  \begin{tabular}{@{}lcccc@{}}
    \toprule
     & \textbf{CD~$\downarrow$} & \textbf{PSNR~$\uparrow$} & \textbf{SSIM~$\uparrow$} \\
    \midrule
    No~Charts Encodings & 2.693 & 16.37 & 0.369 \\
    No~Depth Encodings &  \cellcolor{yellow!25}1.601 & \cellcolor{yellow!25}17.38 & \cellcolor{yellow!25}0.424  \\
    \midrule
    No~$\mathcal{L}_{\text{fit}}$ confidence & 1.703 & \cellcolor{orange!25}17.39 & \cellcolor{orange!25}0.428  \\
    No~$\mathcal{L}_{\text{struct}}$ & 1.716 & 17.00 & 0.410  \\
    No~$\mathcal{L}_{\text{align}}$ & \cellcolor{orange!25} 1.565 & 17.33 & \cellcolor{yellow!25}0.424 \\
    \midrule
    Full Model &  \cellcolor{red!25}\textbf{1.04} & \cellcolor{red!25}\textbf{17.59} & \cellcolor{red!25}\textbf{0.443} \\
    \bottomrule
  \end{tabular}
  }
  \caption{\textbf{Ablation studies on the deformation model and loss components.} For evaluating in bounded scenes, we compute the Chamfer Distance on the DTU dataset~\cite{aanaes2016large}. For evaluating in unbounded scenes, we optimize our representation on 5 images of each scene of the Mip-NeRF~360~\cite{barron22mipnerf360} dataset, and we compute rendering metrics on 10 challenging views with little overlap.
  We see that the charts encodings~(\textbf{CE}), the 1D~depth encodings~(\textbf{1D-DE}), and all of the loss components are required for reaching optimal performance.}
  \label{tab:ablation_deformation_and_losses}
\end{table}

\vspace{-8pt}
\paragraph{Qualitative Comparisons with Sparse-View Methods}
\cref{fig:comparison_with_SOTA} shows qualitative comparisons with the state-of-the-art sparse-view surface reconstruction~\cite{raj2024spurfies} and novel view synthesis~\cite{chen2024mvsplat} methods. Existing methods struggle to generalize to the unbounded scene. In contrast, our method achieves high-quality surface reconstruction even for a scene with a complex background.

\vspace{-8pt}
\paragraph{Ablation Studies}
We conduct extensive ablation studies on the deformation architecture and loss components to validate our design choices on the DTU dataset~\cite{aanaes2016large} and Mip-NeRF 360 dataset~\cite{barron22mipnerf360}.
As shown in \cref{tab:ablation_deformation_and_losses}, removing either the charts encodings or 1D depth encodings leads to decreased performance, confirming the importance of both components.
The results also show the effectiveness of the weighting of the fitting loss $\mathcal{L}_{\text{fit}}$ with learnable confidence maps, the structure loss $\mathcal{L}_{\text{struct}}$, and the mutual alignment loss $\mathcal{L}_{\text{align}}$. 
\section{Conclusion}

\guedon{
We presented a novel approach for reconstructing high-quality 3D surface meshes from sparse-view RGB images within minutes. Our method leverages a novel representation that models surfaces as 2D manifolds through a collection of charts, initialized using pretrained monocular depth estimation. By combining geometric priors from a depth estimation model, efficient chart-based deformation, and differentiable rendering with 2D Gaussian surfels, our method is able to extract a sharp and accurate estimate of the 3D scene from a few RGB images only. 
Our representation also allows for high-quality rendering and significantly faster optimization than other state-of-the-art methods.
Our experiments demonstrate that our method outperforms existing approaches in sparse-view scenarios while being significantly faster, typically requiring only a few minutes for complete reconstruction.

While our method shows promising results, there are several directions for future work. Extending our framework to handle dynamic scenes and deformable objects would broaden its applications in computer vision and graphics.
We believe our work opens new possibilities for fast, high-quality 3D reconstruction from sparse views, with potential applications in virtual reality, digital content creation, and robotics.
}

\section*{Acknowledgements}

This work was in part supported by
JSPS 
20H05951 and %
21H04893, %
and JST JPMJCR20G7 %
and JPMJAP2305. 
%
This work was also in part supported by the ERC grant “explorer ” (No. 101097259).
%
This work was granted access to the HPC resources of IDRIS under the allocation 
2024-AD011013387R2 
made by GENCI.

{
    \small
    \bibliographystyle{ieeenat_fullname}
    \bibliography{main}
}

\maketitlesupplementary
\setcounter{page}{1}

In this appendix, we describe 
\begin{itemize}
    \item additional implementation details,
    \item details about our mesh extraction method,
    \item and additional qualitative results.
\end{itemize}
We also provide a \href{https://anttwo.github.io/matcha/}{video} that offers an overview of the approach and showcases additional qualitative results.

\section{Implementation Details}

\subsection{Initializing Charts}

For initializing the charts using a monocular depth estimation model, we not only backproject the depth maps into 3D but also roughly adjust the scale of the depth estimates using a global affine rescaling model~\cite{turkulainen2024dn,kerbl2024hierarchicalgaussians}. 
Note that we can compute an explicit closed-form solution for this affine rescaling which executes in less than a second. 

In our experiments, the size of our charts is proportional to the input views, and the longest sides of the charts have length $\max(h, w) = 512$. 
We rely on MASt3R-SfM~\cite{duisterhof2024mast3rsfm} to obtain an SfM point cloud for aligning the charts. 

\subsection{Chart Deformation Model} 

We can adjust the resolution of the learnable charts encodings (\ie, $r$) according to the density of the SfM points or the number of views. The sparser the SfM point cloud or the training images, the lower the resolution of the charts encodings. In other words, we can explicitly adjust the strength of the inductive bias in our chart deformation model according to the different scenarios. 
For small scenes with only 3 input views like the objects from the DTU~\cite{aanaes2016large} dataset, we use a small resolution parameter $r=0.1$ for the charts encodings.
In larger and unbounded scenes with 5 or 10 input views, we use a larger resolution parameter $r=0.4$ for our charts encodings.

The other hyperparameters are constants and independent of the inputs. In practice, we set $d=32$ and use an MLP with only 1 hidden layer. The number of channels in the hidden layer is 64. 
For aligning our charts with the initial SfM points, we optimize our model for 1000 iterations. For refining the charts, we optimize our model for 3000 iterations.

During the alignment with the SfM points, we deform the charts along the camera rays, as we empirically found it to be more robust. Moreover, deforming the charts along the camera rays enables very efficient computation of the mutual alignment loss, as in this case, the 3D to 2D mapping of our charts is equivalent to the camera screen projection transform.
To deform charts along the rays, we use a one-dimensional output layer for the MLP, and we compute the 3D deformation by multiplying the MLP output by the ray direction. 

During the refinement with Gaussian surfel rendering, we first update the initial charts $\psi_i^{(0)}$ and replace them with the deformed charts $\psi_i$; Then, we reinitialize the weights of the MLP and replace the output layer with a 3-dimensional layer in order to learn a full 3D deformation for the charts.

\subsection{Refining the Manifold with Gaussian Surfels}

During the second optimization stage, we rely on a photometric loss to refine the manifold. 
At each iteration, we render the manifold by first instantiating 2D Gaussian surfels on the surface, then rasterizing the Gaussians with a surfel rasterizer~\cite{huang20242d}.

\paragraph{Photometric loss} The photometric loss consists of an L1 loss $\mathcal{L}_1$ and a D-SSIM term $\mathcal{L}_{\text{D-SSIM}}$:
\begin{equation}
    \mathcal{L}_{\text{photo}} = (1-\lambda)\mathcal{L}_1 + \lambda\mathcal{L}_{\text{D-SSIM}}\,,
\end{equation}
where we set $\lambda=0.2$, following past 3DGS works~\cite{kerbl3Dgaussians}.

\paragraph{Structure loss} To preserve the fine geometry of our aligned charts, we maintain the structure loss but replace the depth estimates with the depth of our aligned charts. We also weight the structure loss using our confidence maps $C_i$ estimated during the manifold alignment to the SfM points, which makes it a scale-accurate depth regularization robust to outliers.

To further regularize the geometry, we also use a depth-normal consistency loss and a depth distortion loss, as introduced in 2DGS~\cite{huang20242d}.

\paragraph{Distortion loss} The distortion loss prevents Gaussians from spreading around the surface. Since we instantiate Gaussian surfels on the manifold represented as a collection of charts, the distortion term enforces the surfaces of the different charts to align together and form a coherent manifold. 
For each pixel $p$, the distortion loss is given by
\begin{equation}
    \mathcal{L}_{d} = \sum_{i,j}\omega_i \omega_j |z_i - z_j| \,,
\end{equation}
where $i$ and $j$ represent the $i$-th and $j$-th Gaussian surfels intersected along the ray, $z_i$ is the depth of the intersection point between the ray and the $i$-th Gaussian surfel, and $\omega_i$ is the blending weight of the $i$-th intersection.

\paragraph{Depth-Normal consistency loss} The depth-normal consistency loss aims to align the normals of the closest Gaussian surfels along the ray with the gradient of the depth map. In our case, this term encourages the surfaces of the different charts to have the same orientation. 
For a pixel $p$, the normal consistency loss at $p$ is given by
\begin{equation}
    \mathcal{L}_{n} = \sum_{i}\omega_i (1 - \mathbf{n}_i^\mathrm{T}\mathbf{N}_p)\,,
\end{equation}
where $\mathbf{n}_i$ is the normal of the $i$-th Gaussian surfel along the ray and $\mathbf{N}_p$ is the normal at pixel $p$ computed from the gradient of the depth map.

\paragraph{Optimization loss} The complete loss for refining the charts is
\begin{equation}
    \mathcal{L}_{\text{refine}} = \mathcal{L}_{\text{photo}} + \lambda_{\text{struct}}\mathcal{L}_{\text{struct}} +
    \lambda_{d}\mathcal{L}_{d} + \lambda_n \mathcal{L}_{n}\,,
\end{equation}
where we set $\lambda_{\text{struct}} = 1$, $\lambda_{d} = 500$, and $\lambda_{n} = 0.25$. We refine the representation for 3000 iterations, and introduce $\mathcal{L}_d$ and $\mathcal{L}_n$ only after 600 iterations. \koheirmk{The total number of iterations is already introduced in Sec. 1.2. So we may be able to simplify this, \eg, We introduce $\mathcal{L}_d$ and $\mathcal{L}_n$ only after 600 out of the 3000 iterations (I'm not sure if this is good English.).}

\subsection{Extracting a Surface Mesh from the Manifold}

We propose two different approaches for extracting a surface mesh from our manifold representation, depending on the scene complexity and the desired level of detail.

\paragraph{Direct Mesh Extraction} For scenes with moderate complexity or extreme sparse-view setups (e.g., 3 views on DTU), we can directly extract a surface mesh from our manifold using a custom multi-resolution TSDF fusion approach, or a custom implementation of the adaptive tetrahedralization from Gaussian~Opacity~Fields~\cite{yu2024gaussian}. Since we describe our tetrahedralization in the main paper, we focus on providing additional details about the multi-resolution TSDF below.

We render depth maps from our manifold and fuse them into several TSDF volumes with different resolutions. The lower the resolution, the larger the bounding box used for applying the TSDF algorithm.
Then, we merge the TSDF volumes and remove the overlapping regions. Note that, in a sparse-view scenario, the number of depth maps is very low, so that integrating depth maps for computing the TSDF volumes is very fast and takes less than a minute.

Our multi-resolution approach allows us to accurately reconstruct both foreground objects and background regions with a decent number of vertices, which is crucial for unbounded scenes. 
However, even though our multi-resolution TSDF is very fast, it generally erodes the geometry and creates holes in the extracted surface. In this regard, we recommend using the tetrahedralization for extracting meshes.

\paragraph{Free Gaussians Refinement} For scenes requiring finer geometric details, particularly in large unbounded environments, we propose an additional refinement step that leverages our manifold as a strong geometric prior. Instead of directly extracting the mesh, we first let Gaussian surfels get freely optimized in 3D space for a few iterations while strongly constraining them with our manifold representation. 

For this, we freeze the manifold but unfreeze the Gaussians' parameters (position, scale, and rotation) and regularize them using depth maps rendered from the manifold through a combination of our refinement loss and an L1 depth loss with a weighting factor $\lambda_{depth}=0.75$. We also use our confidence maps to weigh the depth regularization, but not the structure loss that relies on the derivatives of the depth. Indeed, applying the structure loss everywhere in the scene enables regularization of Gaussians located even in low-confidence areas, where normal maps and curvature maps still provide a reliable supervision signal despite of inaccurate depth values.

This refinement step is particularly effective because our manifold provides scale-accurate regularization, unlike traditional depth-based regularization methods that often struggle with scale ambiguity. The manifold acts as a reliable geometric prior that prevents Gaussians from diverging while letting them recover fine surface details that might not be fully captured by the manifold representation alone.

After this Gaussian refinement stage, we extract the final mesh using the same multi-resolution TSDF fusion or tetrahedralization approaches described above, but now applied to the refined Free Gaussians representation. This two-stage approach allows us to recover very fine geometric details while maintaining the overall accuracy and robustness of our manifold representation.

\section{Additional Results and Details}

\begin{figure*}[t]
    \centering
    \begin{minipage}[b]{0.02\textwidth}
        \centering
        \rotatebox{90}{3 views}
        \vspace{45pt} 
        \vfill
        \rotatebox{90}{3 views}
        \vspace{40pt} 
        \vfill
        \rotatebox{90}{5 views}
        \vspace{40pt} 
        \vfill
        \rotatebox{90}{5 views}
        \vspace{35pt} 
        \vfill
        \rotatebox{90}{10 views}
        \vspace{40pt} 
        \vfill
        \rotatebox{90}{10 views}
        \vspace{40pt} 
        \vfill
        \rotatebox{90}{10 views}
        \vspace{20pt} 
        \vfill
    \end{minipage}
    \begin{minipage}[b]{0.24\textwidth}
        \centering
        Rendering
        \vspace{2pt}
        \vfill
        \includegraphics[trim={0 0cm 0 0cm},clip,width=0.98\linewidth]{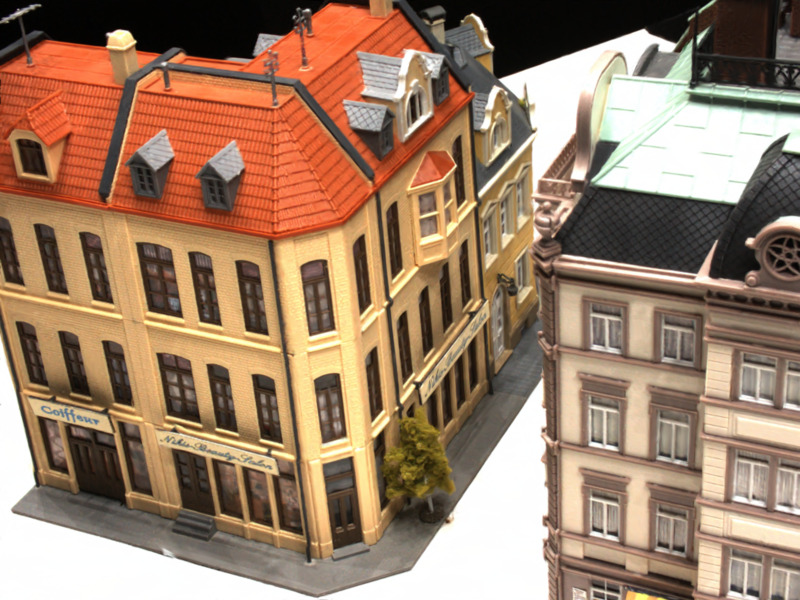}
        \includegraphics[trim={0 0cm 0 7.65cm},clip,width=0.98\linewidth]{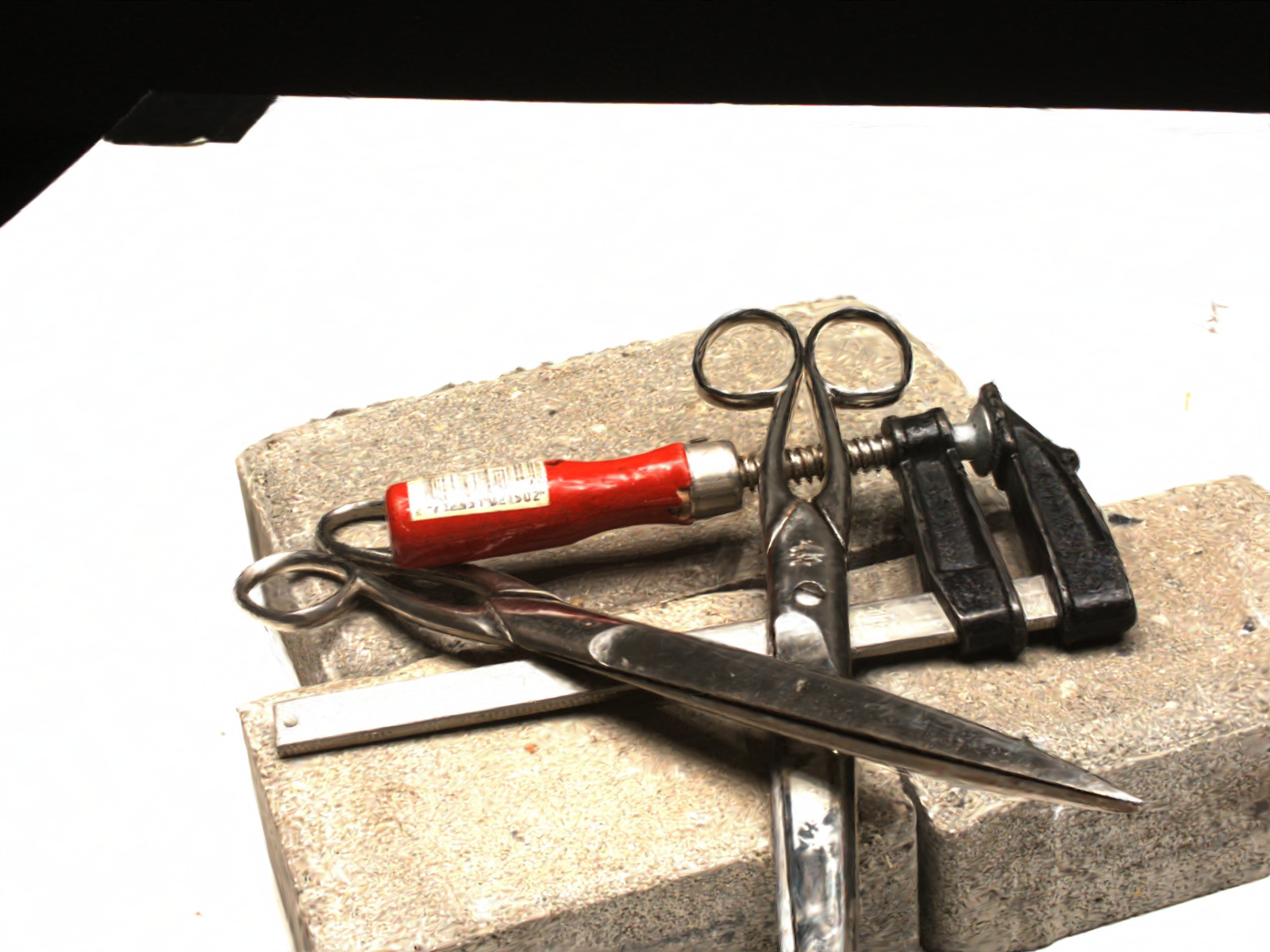}
        \includegraphics[trim={0 0cm 0 0cm},clip,width=0.98\linewidth]{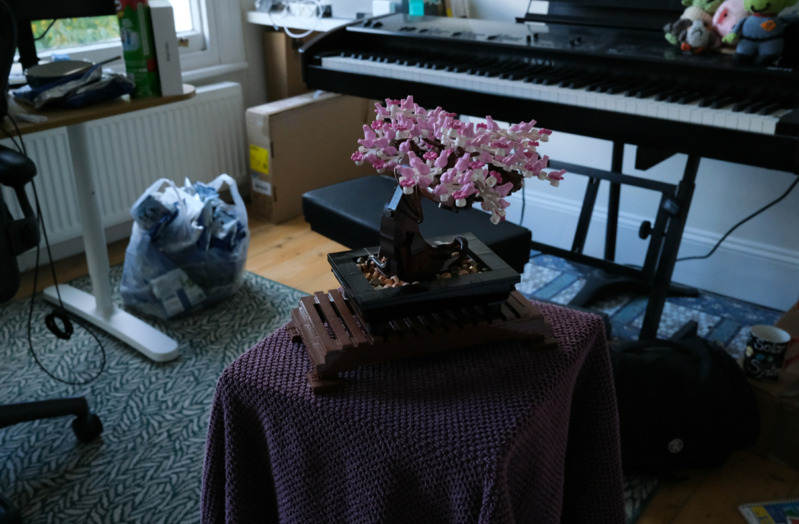}
        \includegraphics[trim={0 0cm 0 0cm},clip,width=0.98\linewidth]{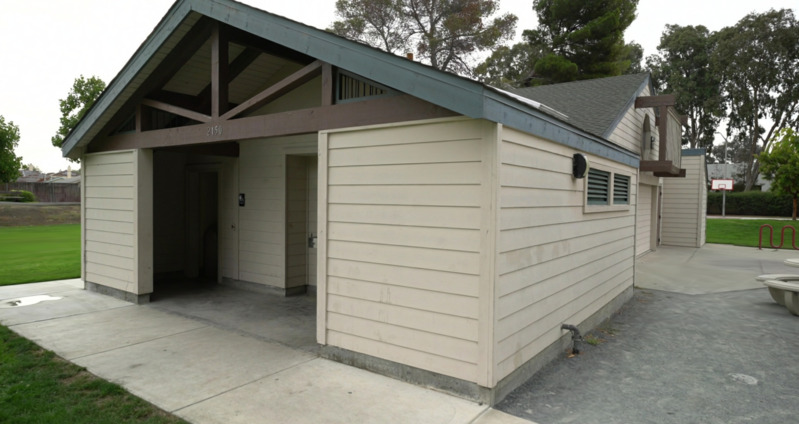}
        \includegraphics[trim={0 0cm 0 0cm},clip,width=0.98\linewidth]{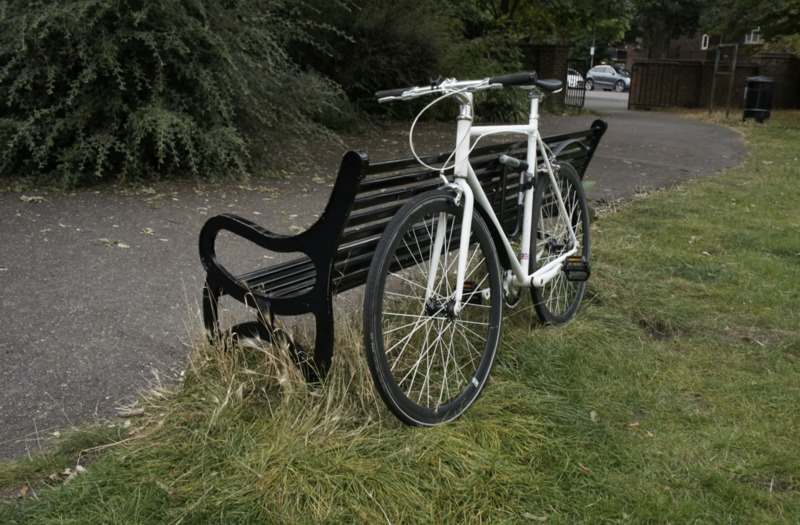}
        \includegraphics[trim={0 0cm 0 0cm},clip,width=0.98\linewidth]{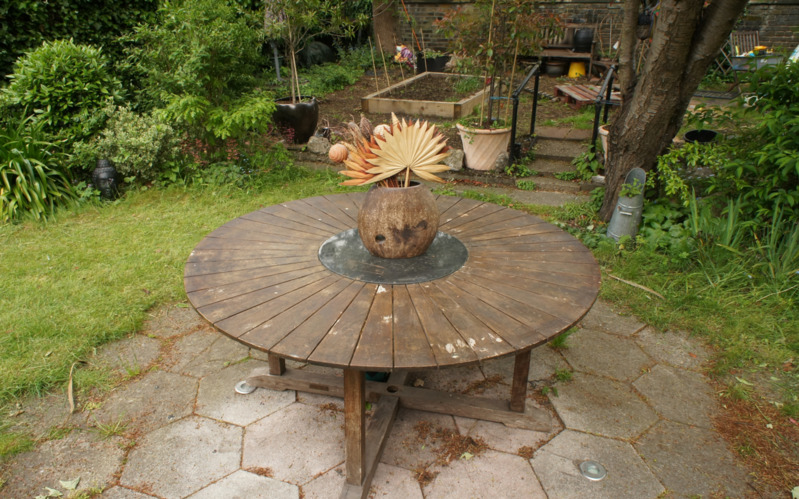}
        \includegraphics[trim={0 0cm 0 0cm},clip,width=0.98\linewidth]{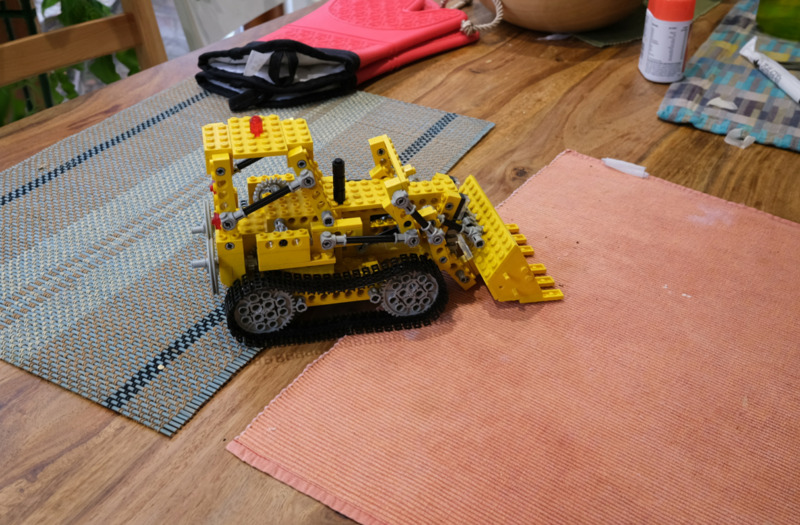}
    \end{minipage}
    \begin{minipage}[b]{0.24\textwidth}
        \centering
        Depth
        \vspace{2pt}
        \vfill
        \includegraphics[trim={0 0cm 0 0cm},clip,width=0.98\linewidth]{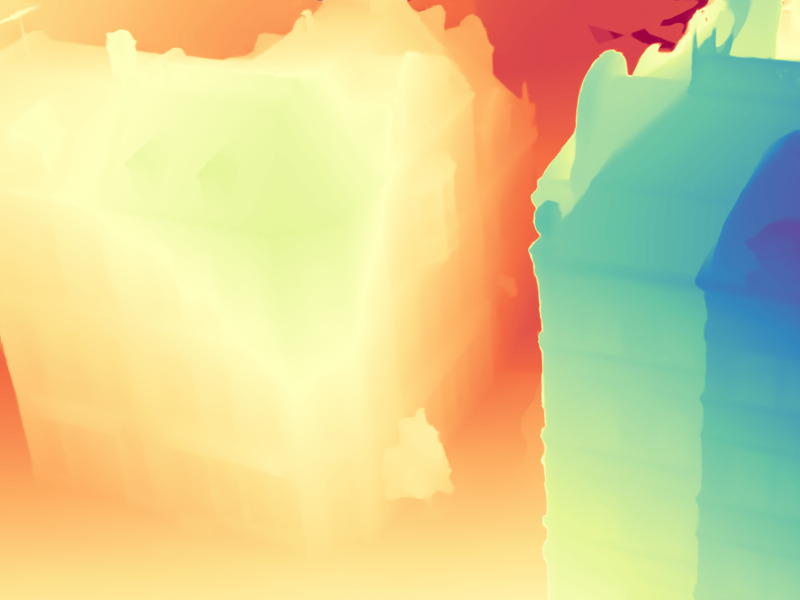}
        \includegraphics[trim={0 0cm 0 7.65cm},clip,width=0.98\linewidth]{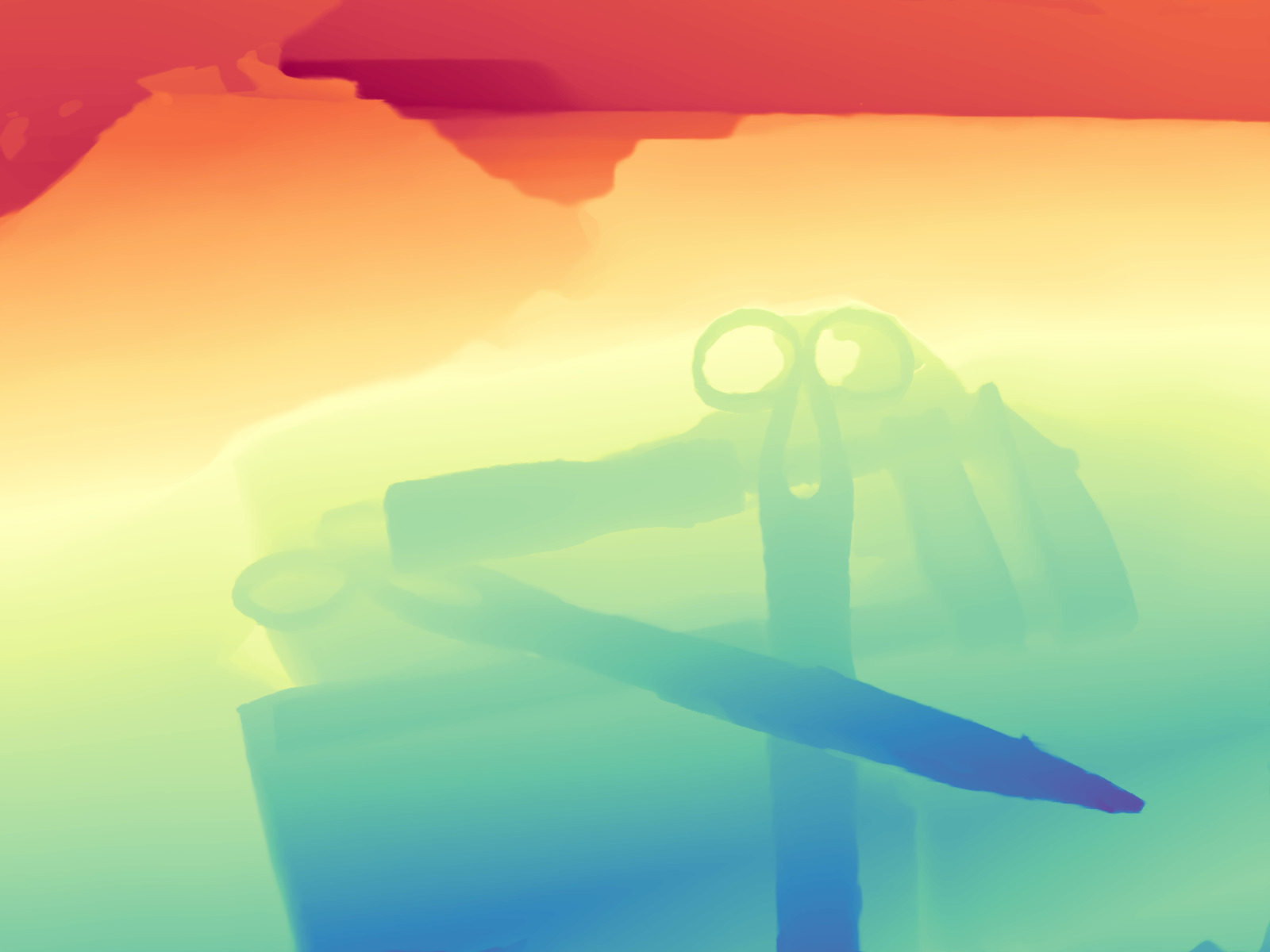}
        \includegraphics[trim={0 0cm 0 0cm},clip,width=0.98\linewidth]{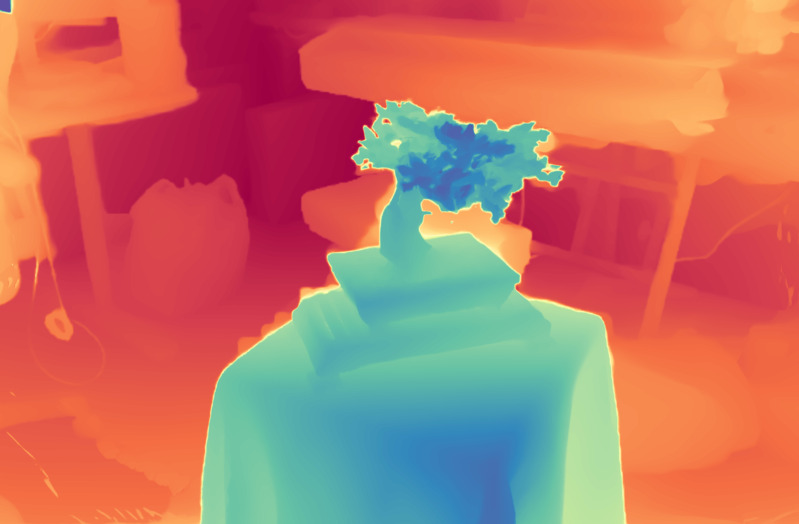}
        \includegraphics[trim={0 0cm 0 0cm},clip,width=0.98\linewidth]{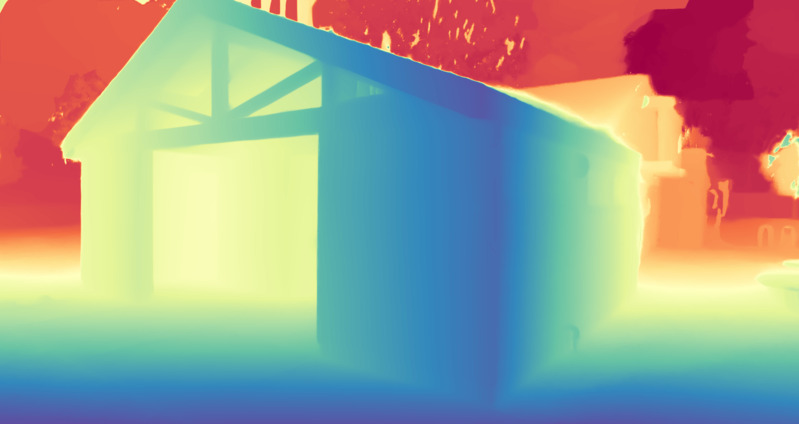}
        \includegraphics[trim={0 0cm 0 0cm},clip,width=0.98\linewidth]{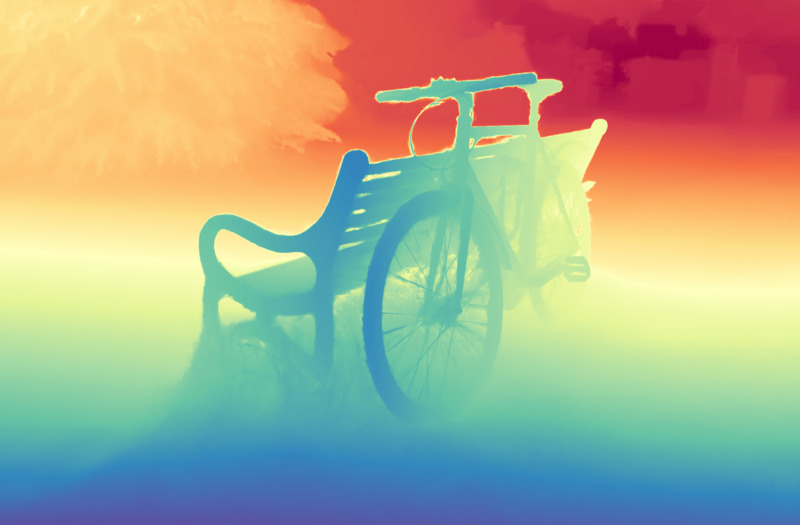}
        \includegraphics[trim={0 0cm 0 0cm},clip,width=0.98\linewidth]{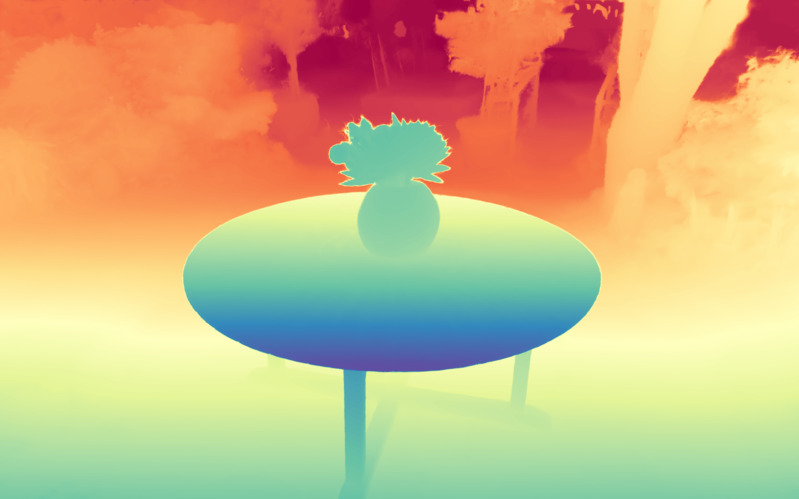}
        \includegraphics[trim={0 0cm 0 0cm},clip,width=0.98\linewidth]{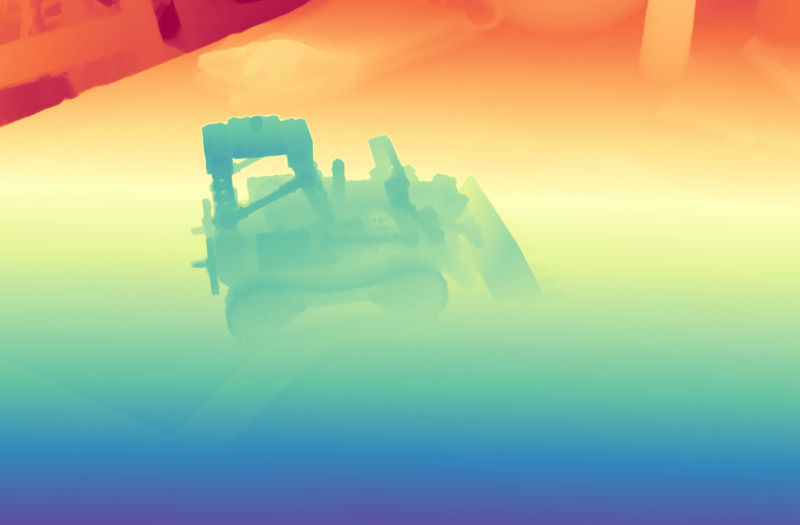}
    \end{minipage}
    \begin{minipage}[b]{0.24\textwidth}
        \centering
        Normal
        \vspace{4pt}
        \vfill
        \includegraphics[trim={0 0cm 0 0cm},clip,width=0.98\linewidth]{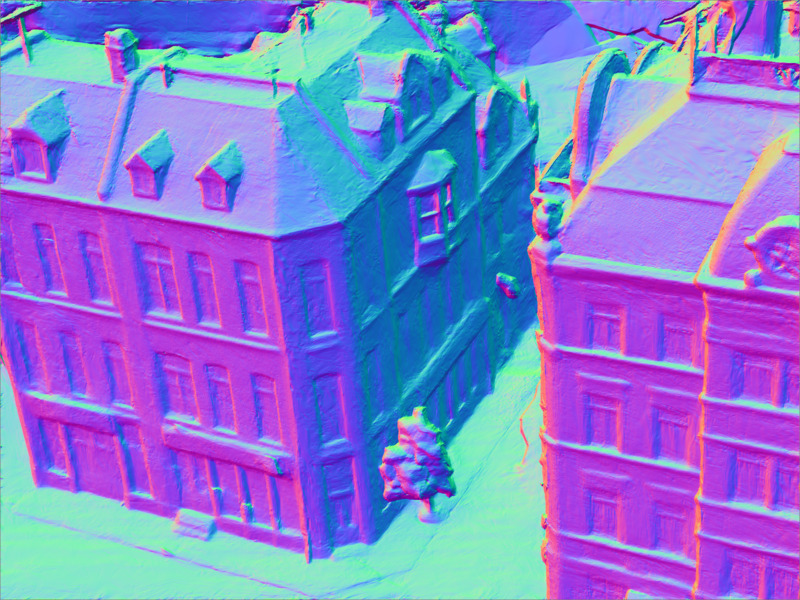}
        \includegraphics[trim={0 0cm 0 7.65cm},clip,width=0.98\linewidth]{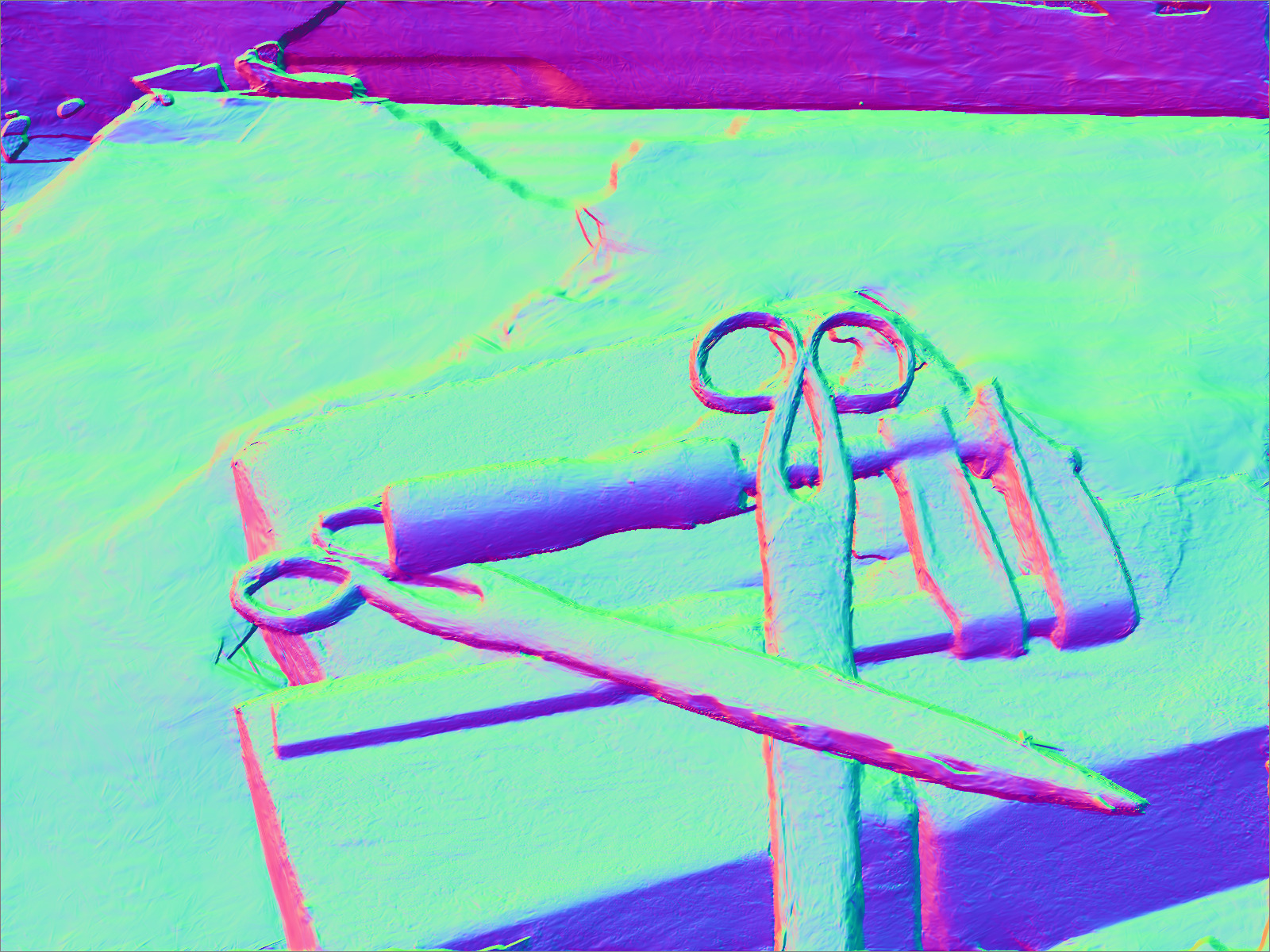}
        \includegraphics[trim={0 0cm 0 0cm},clip,width=0.98\linewidth]{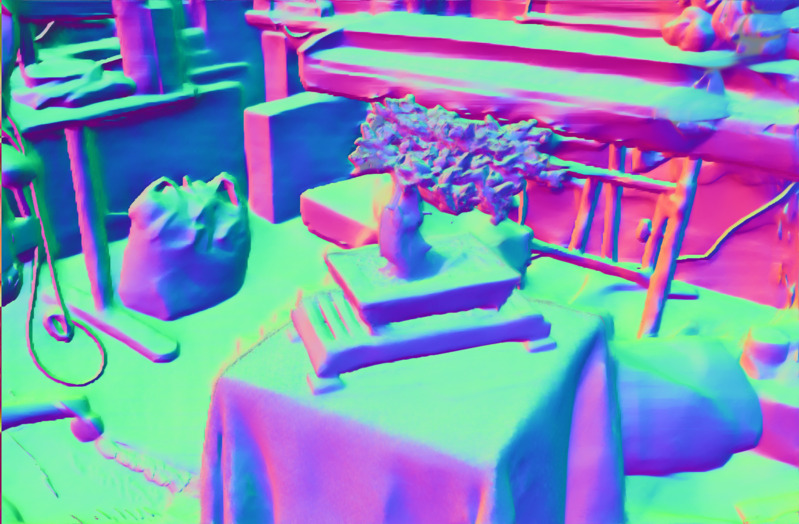}
        \includegraphics[trim={0 0cm 0 0cm},clip,width=0.98\linewidth]{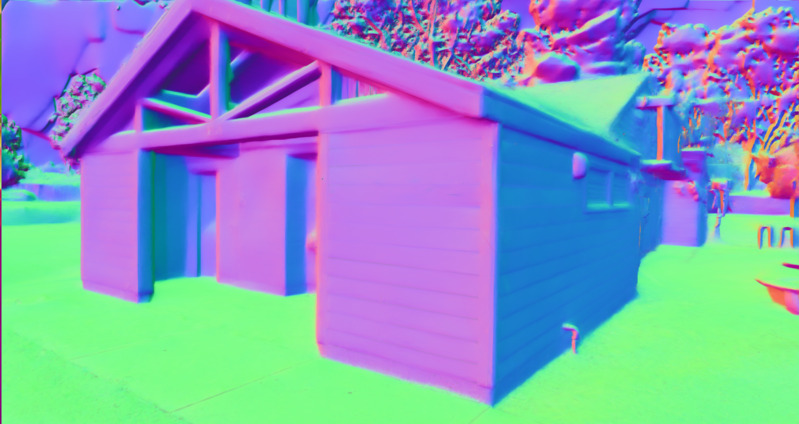}
        \includegraphics[trim={0 0cm 0 0cm},clip,width=0.98\linewidth]{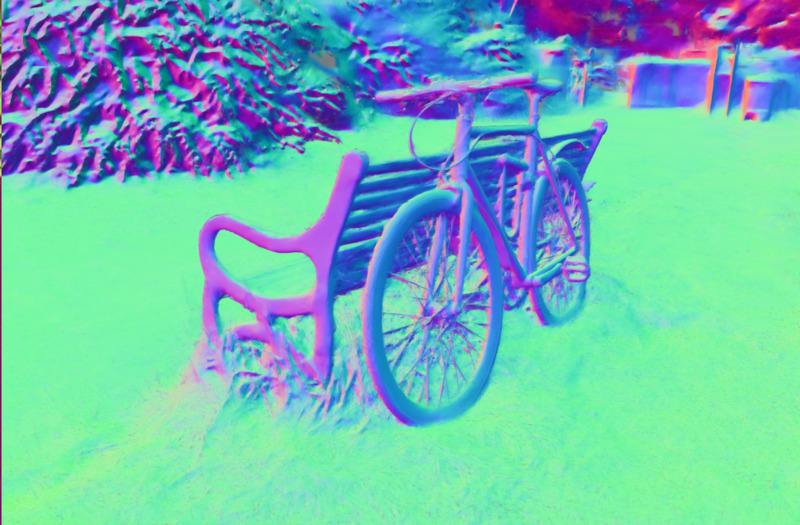}
        \includegraphics[trim={0 0cm 0 0cm},clip,width=0.98\linewidth]{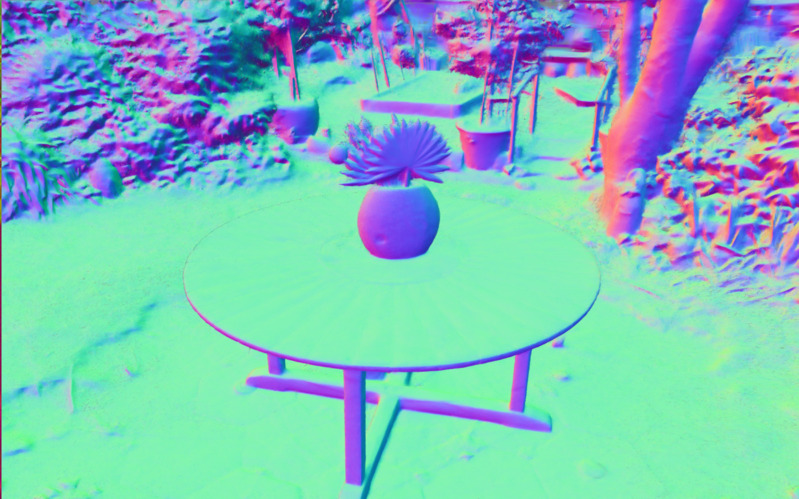}
        \includegraphics[trim={0 0cm 0 0cm},clip,width=0.98\linewidth]{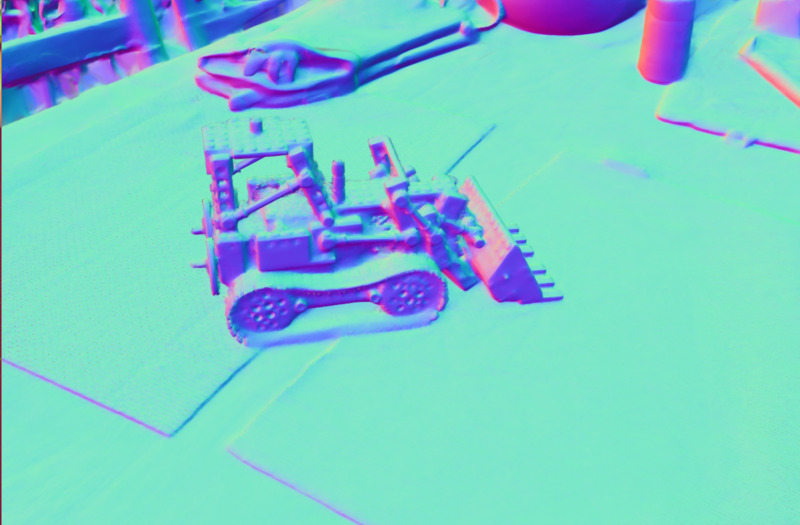}
    \end{minipage}
    \begin{minipage}[b]{0.24\textwidth}
        \centering
        Mesh
        \vspace{4pt}
        \vfill
        \includegraphics[trim={8.5cm 0cm 8.5cm 0cm},clip,width=0.98\linewidth]{images/mesh/scan21_1.jpg}
        \includegraphics[trim={0 0cm 0 0cm},clip,width=0.98\linewidth]{images/mesh/scan37_1.jpg}
        \includegraphics[trim={1.75cm 0cm 8cm 0cm},clip,width=0.98\linewidth]{images/mesh/bonsai_5_views_0.jpg}
        \includegraphics[trim={0 0cm 0 2cm},clip,width=0.98\linewidth]{images/mesh/barn_5_views.jpg}
        \includegraphics[trim={0 0cm 9.5cm 0cm},clip,width=0.98\linewidth]{images/mesh_extraction/adaptive_tetra.jpg}
        \includegraphics[trim={0 0cm 6.2cm 0cm},clip,width=0.98\linewidth]{images/mesh/garden_10_views_0.jpg}
        \includegraphics[trim={10cm 0cm 0 0cm},clip,width=0.98\linewidth]{images/mesh/kitchen_10_views_1.jpg}
    \end{minipage}
    
    \caption{\textbf{Qualitative reconstruction results across different scenarios and numbers of input views.} We show results on both bounded objects from DTU~\cite{aanaes2016large} (first two rows, 3 views) and unbounded scenes from Tanks\&Temples~\cite{Knapitsch2017} and Mip-NeRF 360~\cite{barron22mipnerf360} (middle and bottom rows). For each example, we show (from left to right): the rendered novel view, estimated depth map, surface normals, and the extracted mesh. For bounded objects (DTU), meshes are extracted directly from our manifold representation, while for unbounded scenes, we first refine free Gaussians around the manifold before mesh extraction. Note how our method maintains consistent quality across different scenarios, from small objects to large-scale scenes with complex backgrounds.}
    \label{fig:additional qualitative results}
\end{figure*}

\tocheck{\begin{figure*}[t]
    \centering
    \begin{minipage}[b]{0.32\textwidth}
        \centering
        2DGS~\cite{huang20242d} + MASt3R-SfM~\cite{duisterhof2024mast3rsfm}
        \vspace{2pt}
        \vfill
        \includegraphics[trim={0 0cm 0 0cm},clip,width=0.98\linewidth]{example-image-a}
        \includegraphics[trim={0 0cm 0 0cm},clip,width=0.98\linewidth]{example-image-b}
    \end{minipage}
    \begin{minipage}[b]{0.32\textwidth}
        \centering
        GOF~\cite{yu2024gaussian} + MASt3R-SfM~\cite{duisterhof2024mast3rsfm}
        \vspace{2pt}
        \vfill
        \includegraphics[trim={0 0cm 0 0cm},clip,width=0.98\linewidth]{example-image-a}
        \includegraphics[trim={0 0cm 0 0cm},clip,width=0.98\linewidth]{example-image-b}
    \end{minipage}
    \begin{minipage}[b]{0.32\textwidth}
        \centering
        \textbf{\method~(Ours)}
        \vspace{2pt}
        \vfill
        \includegraphics[trim={0 0cm 0 0cm},clip,width=0.98\linewidth]{example-image-a}
        \includegraphics[trim={0 0cm 0 0cm},clip,width=0.98\linewidth]{example-image-b}
    \end{minipage}
    
    \caption{\textbf{Qualitative comparisons of novel view synthesis.} \tocheck{Add results and captions or remove this. If added, do not forget to remove tocheck macro around the input command in X\_supp.tex.}}
    \label{fig:additional qualitative comparison NVS}
\end{figure*}

}

\paragraph{Surface Reconstruction}
\cref{fig:additional qualitative results} shows qualitative results. Our method can reconstruct high-quality surfaces across different scenarios, from bounded objects (DTU~\cite{aanaes2016large} dataset) to unbounded scenes (Tanks\&Temples~\cite{Knapitsch2017} and Mip-NeRF~360~\cite{barron22mipnerf360} datasets), using varying numbers of, but sparse, input views (3, 5, and 10 views).
For each example, we show a rendered view, the estimated depth map, surface normals, and the extracted mesh, which collectively show the consistency of our reconstruction across different representations. For the objects from the DTU dataset, we directly extract the mesh from the manifold representation using our multi-resolution TSDF fusion approach. For the unbounded scenes from the T\&T and Mip-NeRF 360 datasets, we first refined free Gaussians around the manifold as explained in the previous section, then extracted the mesh using the same multi-resolution TSDF fusion approach.
\koheirmk{The details of the mesh extraction are in the section right before this section. So these might be redundant.}

In the 3-view scenarios (first two rows), our method successfully recovers detailed geometry despite the extreme sparsity of input views. The 5-view and 10-view examples (middle and bottom rows) demonstrate how our approach scales to larger unbounded scenes while maintaining reconstruction quality throughout the scene, including distant background regions.

\begin{figure}[t]
    \centering
    \begin{minipage}[b]{0.23\textwidth}
        \centering
        MVSplat360~\cite{chen2024mvsplat360}
        \vspace{2pt}
        \vfill
        \includegraphics[width=1.\linewidth]{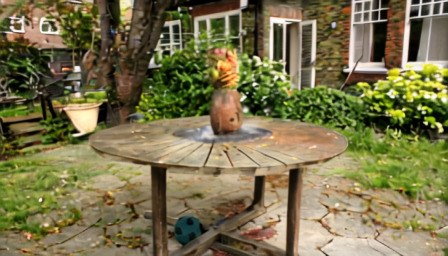}
    \end{minipage}
    \begin{minipage}[b]{0.23\textwidth}
        \centering
        Ours
        \vspace{4pt}
        \vfill
        \includegraphics[trim={1.5cm 2.4cm 1.4cm 2.4cm},clip,width=1.\linewidth]{images/rendering/garden_5_views.jpg}
    \end{minipage}
    
    \caption{\textbf{Qualitative comparisons of novel view synthesis with MVSplat360~\cite{chen2024mvsplat360} on an unbounded scene with 5 training images.} 
    Our method can render more photorealistic images than the concurrent feed-forward novel view synthesis method in sparse view scenarios.}
    \label{fig:comparison_with_mvsplat360}
\end{figure}

\paragraph{Novel View Synthesis}
\cref{tab:mipzipdb_nvs} of the main paper provides results of novel view synthesis in sparse-view settings across three challenging real-world datasets of unbounded scenes: Mip-NeRF 360~\cite{barron22mipnerf360}, Tanks\&Temples~\cite{Knapitsch2017}, and DeepBlending~\cite{hedman18deepblending}.
Specifically, we follow \kohei{3DGS}~\cite{kerbl3Dgaussians} and use the scenes \textit{Playroom} and \textit{Dr.~Johnson} for evaluation on the DeepBlending dataset. For the T\&T dataset, we use the standard split of 6 scenes as used in \kohei{2DGS~\cite{huang20242d} and GOF~\cite{yu2024gaussian}}
but removed \textit{Courthouse} and \textit{Meetingroom}, as these very large scenes are not suitable for sparse-view scenarios with only 5 or 10 input views.

We consider two scenarios with 5 and 10 training views, respectively. For each dataset, we built training sets of 5 and 10 input views and evaluated on a set of 10 test views, including both easy views with high overlap with the training views and much more challenging views with very limited overlap. For fair comparison, we augment recent state-of-the-art methods (2DGS~\cite{huang20242d} and GOF~\cite{yu2024gaussian}) with MASt3R-SfM~\cite{duisterhof2024mast3rsfm}, as it provides better robustness in sparse-view scenarios.

We report both average PSNR and 10\% quantile PSNR (10\%Q PSNR) metrics. 
\kohei{The 10\%Q PSNR is the PSNR value below which 10\% of the test views fall. Average PSNR provides an overall measure of reconstruction quality. In contrast, the 10\%Q PSNR specifically captures accuracy on the most challenging views as well as the ability of a method to generalize to novel viewpoints. }
This metric is particularly relevant to sparse-view settings where some novel viewpoints may have very limited overlap with input views.

As shown in \cref{tab:mipzipdb_nvs}, our method consistently outperforms the baselines across all datasets and metrics. In the 5-view scenario, we achieve significant improvements over the baselines. The performance gap remains substantial even when increasing to 10 input views, where our method maintains superior reconstruction quality across datasets. This consistent performance advantage demonstrates the effectiveness of our chart-based representation and refinement approach in handling sparse-view scenarios.

Notably, our method shows particular strength in maintaining quality for challenging views, as evident in the larger improvements in 10\%Q PSNR compared to average PSNR. This suggests that our chart-based representation, combined with the robust deformation model and multi-stage refinement process, helps maintain consistency even in regions with limited overlap in views.

\tocheck{\cref{fig:additional qualitative comparison NVS} shows qualitative comparisons of novel view synthesis with the baseline methods. Although the baselines cause artifacts in the rendered images, our method realizes high-quality and photorealistic rendering without artifacts.}

We also qualitatively compare our method with MVSplat360~\cite{chen2024mvsplat360}, a concurrent method for feed-forward novel view synthesis in sparse-view settings. \cref{fig:comparison_with_mvsplat360} shows rendering results of MVSplat360 and our method from novel viewpoints by using 5 views. MVSplat360 suffers from a domain gap from training data and limited image resolution due to training of its feed-forward networks, leading to unrealistic rendering results.

\end{document}